\documentclass[runningheads]{llncs}

\usepackage{eccv}

\usepackage{eccvabbrv}

\usepackage{graphicx}
\graphicspath{{./}{Supplementary_Material/}}
\usepackage{booktabs}

\usepackage[accsupp]{axessibility}  

\usepackage{hyperref}

\usepackage{orcidlink}

\usepackage{svg}
\usepackage{multirow}
\usepackage[table]{xcolor}
\usepackage{nicematrix}
\usepackage{subcaption} 
\usepackage{placeins}

\definecolor{BrightPearl}{HTML}{F0EEF9}
\definecolor{SoftLavender}{HTML}{E8E6F1}
\definecolor{UltraViolet}{HTML}{F5F3FA}
\definecolor{MistyLavender}{HTML}{E8E6F5}
\definecolor{SweetLavender}{HTML}{E1DDF0}
\definecolor{LavenderGhost}{HTML}{F5F4FB}
\definecolor{LightGreen}{HTML}{cce5d1}
\definecolor{UltraLightGreen}{HTML}{ecf6ef}

\begin{document}

\title{PriorEye: Geospatial Visual Priors for End-to-End Autonomous Driving} 

\titlerunning{PriorEye}

\author{Kyuhwan Yeon\orcidlink{0000-0002-4833-6676} \and
Benjamin Ramtoula\orcidlink{0000-0002-9185-7995} \and
Daniele De Martini\orcidlink{0000-0001-6121-5839}}

\authorrunning{K.~Yeon et al.}

\institute{Mobile Robotics Group, University of Oxford, United Kingdom\\
\email{\{kyuhwan, benjamin, daniele\}@robots.ox.ac.uk}\\
}

\maketitle

\begin{abstract}
    Most end-to-end autonomous driving methods rely solely on instantaneous sensor observations, limiting them to reactive behavior without the anticipatory foresight human drivers employ through prior experience. We introduce geospatial visual priors, street-level visual context anchored to the intended driving route, providing visual-spatial foresight independent of real-time sensors. We propose a memory augmentation module featuring a dual-memory architecture and an adaptive memory gate, which can be easily integrated into existing end-to-end approaches. This design pairs a contextual memory for retrieved priors with a persistent fallback memory, and dynamically regulates the influence of memories based on current state compatibility. Evaluated on the NAVSIM-v2 benchmark, our approach consistently improves performance across diverse end-to-end baselines. Furthermore, because these priors are independent of onboard sensors, our method inherently improves robustness against sensor corruption, while the dual-memory design ensures safe fallback when the retrieved priors themselves become unreliable. Our project page is available at \url{https://ori-mrg.github.io/PriorEye}.
\end{abstract}

\section{Introduction}
\label{sec:intro}

\begin{figure}[t]
  \centering
    \includegraphics[width=\linewidth]{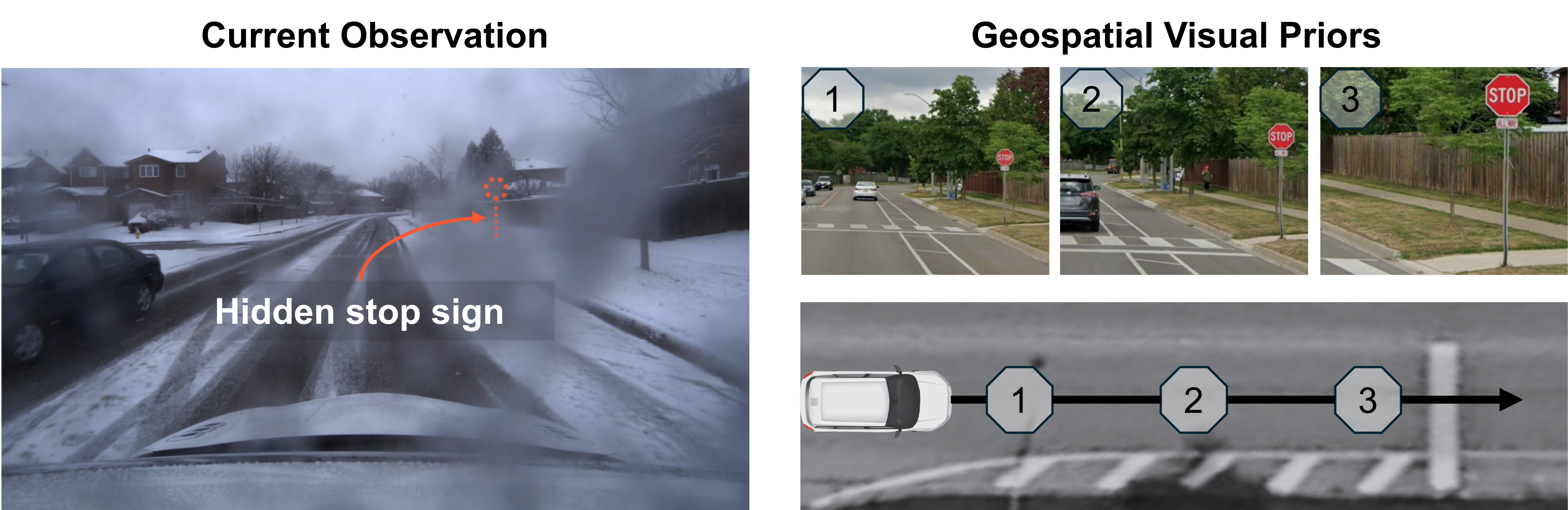}
    \caption{\textbf{Motivation for geospatial visual priors.} \textit{Left:} A current onboard observation from the Boreas dataset~\cite{boreas}, where the camera is occluded by snow and critical features such as stop signs become invisible. Methods that rely solely on instantaneous observations are vulnerable in such conditions. \textit{Right:} Geospatial visual priors, street-level images collected offline along the intended route. Because these priors are independent of current sensor conditions, the stop sign remains clearly visible in the retrieved imagery, enabling anticipatory driving even when onboard perception fails.}
  \label{fig:overall_comparison}
\end{figure}

End-to-end (E2E) autonomous driving has emerged as a compelling paradigm due to its scalability and conceptual simplicity, directly mapping raw sensor observations to driving actions or trajectories~\cite{chenEndtoEndAutonomousDriving2024, tampuuSurveyEndtoEndDriving2022}. By enabling joint optimization and avoiding hand-crafted intermediate representations, E2E approaches mitigate cascading errors inherent to traditional modular pipelines~\cite{Hu_2023_CVPR, Jiang_2023_ICCV, sunSparseDriveEndtoEndAutonomous2025, jiang2026vadv, chittaTransFuserImitationTransformerBased2023, DBLP:conf/eccv/ZhengSGZC24, liuGuideFlowConstraintGuidedFlow2025, kirbyDrivingRegisters2026, Song_2025_CVPR, wu2022trajectory}. Recent advances in vision-language models have further improved performance and enhanced semantic interpretability~\cite{Marcu2023LingoQAVQ, hwang2025emma, nvidiaAlpamayoR1BridgingReasoning2026, Jiang_2025_ICCV}, establishing E2E driving as a prominent approach for autonomous driving.


Despite these advances, a fundamental limitation remains: most existing E2E driving policies are short-horizon reactive, relying only on a few seconds of sensor observations as input \cite{hwang2025emma}. However, effective driving requires reasoning over extended time horizons. Human drivers naturally achieve this by utilizing visual and spatial foresight derived from prior experience~\cite{epsteinCognitiveMapHumans2017, maguireNavigationrelatedStructuralChange2000, rosenbaumHaveOftenWalked2004}. For instance, guided by driving memory, a human driver can proactively merge into an adjacent lane well before a lane ends, even when the lane drop is not yet visible. Similarly, they can decelerate in anticipation of a known sharp curve hidden behind obstacles. This prior information remains uncorrupted by severe weather or visual occlusions. In contrast, current E2E systems, constrained by their exclusive focus on instantaneous sensor observations, lack such anticipatory behavior in complex scenarios. Moreover, this over-reliance on real-time visual cues renders them highly susceptible to sensor corruption, as they lack the persistent contextual evidence necessary to stabilize perception against temporary disturbances.

Inspired by these observations, we propose a mechanism to augment existing end-to-end driving solutions with geospatial visual priors. Acting as a form of driving memory, these priors explicitly encode visual and spatial foresight. In our formulation, spatial priors represent the intended driving route and its associated spatial anchors along that route, while visual priors correspond to offline-collected street-level images that capture the visual context of the driving environment. Geospatial visual priors are constructed by anchoring visual priors to spatial priors. As illustrated in Fig.~\ref{fig:overall_comparison}, these priors provide contextual information about the upcoming driving environment, supplying critical road elements (\eg, stop signs and road markings) that instantaneous onboard sensors may miss due to occlusions or sensor degradation. By coupling street-level visual context with the spatial structure of the intended route, the proposed design enables anticipatory, efficient, and robust driving even under occlusions and limited visibility.

Inspired by memory-augmented Large Language Models (LLMs)~\cite{behrouz2025titans, NEURIPS2023_ebd82705, wuHumanMemoryAI2025a}, we formulate geospatial visual priors as long-term memory. Specifically, we introduce a model-agnostic memory augmentation framework where these priors constitute the contextual memory, providing the driving policy with extended visual-spatial context beyond the immediate sensor view. To ensure robustness against potential retrieval errors, our design complements this with a persistent memory, acting as a fallback reference. An adaptive memory gate explicitly balances the reliance between these memory sources and real-time observations based on similarity cues. Importantly, the proposed module can be seamlessly integrated into a wide range of E2E planners without modifying their core architectures. Our contributions are threefold:

\begin{itemize}
\item We introduce geospatial visual priors that provide spatially grounded visual foresight for end-to-end driving.

\item We propose a model-agnostic dual memory architecture with an adaptive memory gate that effectively fuses geospatial visual priors while preventing over-reliance on misaligned or corrupted memory.

\item We validate that by leveraging priors that are independent of instantaneous onboard sensor observations, our method significantly improves driving performance across diverse E2E planners and enhances robustness under sensor degradation.
\end{itemize}

\section{Related Work}
\label{sec:related_work}
\subsubsection{End-to-End Autonomous Driving.}
Existing E2E driving methods can be broadly categorized by their planning formulations. Regression-based approaches directly predict future trajectories from sensor inputs, offering simplicity and efficiency~\cite{Casas_2021_CVPR,huSTP3EndtoEndVisionBased2022a, Hu_2023_CVPR,chittaTransFuserImitationTransformerBased2023,jia2025drivetransformer,sunSparseDriveEndtoEndAutonomous2025}. Generative planners, including diffusion-based models, model the distribution of feasible trajectories to improve diversity and robustness~\cite{Liao_2025_CVPR, Xing_2025_CVPR,jiangDiffVLAVisionLanguageGuided2025,liuGuideFlowConstraintGuidedFlow2025}. Trajectory scoring methods employ a learned scorer to select the most suitable trajectory from a set of candidate trajectories, enabling effective handling of multi-modal and non-convex planning uncertainty~\cite{liHydraMDPEndtoendMultimodal2024, liHydraMDPAdvancingEndtoEnd2025, liZTRSZeroImitationEndtoend2025, liGeneralizedTrajectoryScoring2025, yaoDriveSuprimPreciseTrajectory2025, kirbyDrivingRegisters2026}. Recent works further incorporate vision-language models to enhance semantic reasoning~\cite{ding2024hintad, xuDriveGPT4InterpretableEndtoEnd2024a, hwang2025emma, Renz_2025_CVPR, nvidiaAlpamayoR1BridgingReasoning2026, Shao_2024_CVPR, Jiang_2025_ICCV}, and widely explore 3D Gaussian Splatting and world models for data augmentation, model evaluation, and performance improvement~\cite{Li_2025_ICCV, tianSimScaleLearningDrive2025, cao2025pseudosimulation, liThink2DriveEfficientReinforcement2025, gao2024vista, Hassan_2025_CVPR, NEURIPS2024_017761f9, zeng2025futuresightdrive, wangDriveDreamerRealWorldDriveWorld2025}. Despite these advances, a common limitation persists: most E2E methods operate within a short input horizon of only a few seconds, restricting them to purely reactive behavior without the ability to reason about upcoming road context beyond the immediate sensor view.

\subsubsection{Priors for Autonomous Driving.}
Prior knowledge in autonomous driving typically refers to information that is not directly observable from online sensors and is obtained offline or from prior experience. Driving experience or memory has been leveraged for high-level reasoning, enabling models to reuse past knowledge from previously encountered scenarios and improve robustness in complex situations~\cite{yuan2024rag,luoMTRDriveMemoryToolSynergistic2025, wangRADRetrievalAugmentedDecisionMaking2025}. 
Spatial priors such as maps, aerial images, and route information encode road topology and traffic rules, primarily used for upstream tasks like online map construction to provide long-range structural guidance beyond the current field of view~\cite{Xiong_2023_CVPR, jiangPMapNetFarSeeingMap2024, mazumderSatMapRevisitingSatellite2026}.
Visual priors such as street-level images exploit offline visual data to capture appearance-level cues and scene context of driving environments~\cite{jiaSpatialRetrievalAugmented2025}.

Although prior information for autonomous driving has been widely explored, existing methods primarily leverage priors for upstream tasks such as online HD map construction or high-level reasoning. In contrast, the direct integration of priors into E2E planning remains largely underexplored. Furthermore, spatial and visual priors are typically considered in isolation, leaving the planner unable to access visual context along the intended route. To address these gaps, our work introduces geospatial visual priors that couple offline visual context with route-level spatial information and inject them directly into E2E driving policies. As a result, our method explicitly provides visual-spatial foresight derived from priors to the E2E planner.
\section{Methodology}
\label{sec:methodology}

We consider E2E driving systems that receive raw sensor observations, the ego-vehicle state, and a high-level navigational command to produce a planned trajectory, and assume access to a lane connectivity graph of the deployment area.

To enable such systems to leverage geospatial visual priors, we first construct a global offline repository of street-level visual embeddings spanning the target area. For a given task, we retrieve relevant embeddings and anchor them to the intended route to form geospatial visual priors (Section~\ref{sec:geospatial_visual_priors}). 

\begin{figure}[!t] 
    \centering
    \begin{subfigure}[b]{\textwidth}
        \centering
        \includegraphics[width=\linewidth, keepaspectratio]{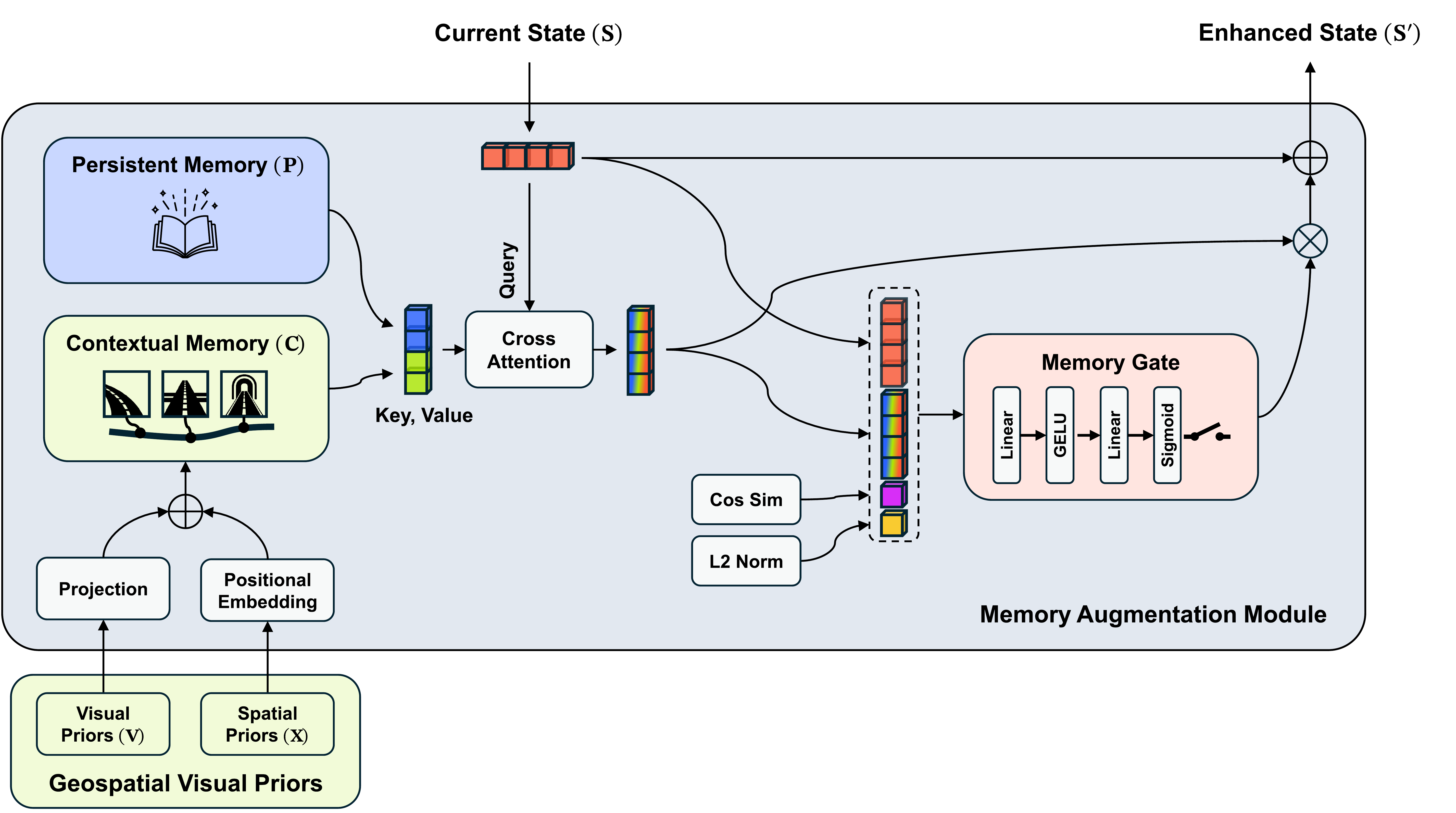}
        \caption{Architecture of the Memory Augmentation Module}
        \label{fig:memory_mechanism_detail}
    \end{subfigure}\\[1em]

    \begin{subfigure}[b]{\textwidth}
        \centering
        \includegraphics[width=\linewidth, keepaspectratio]{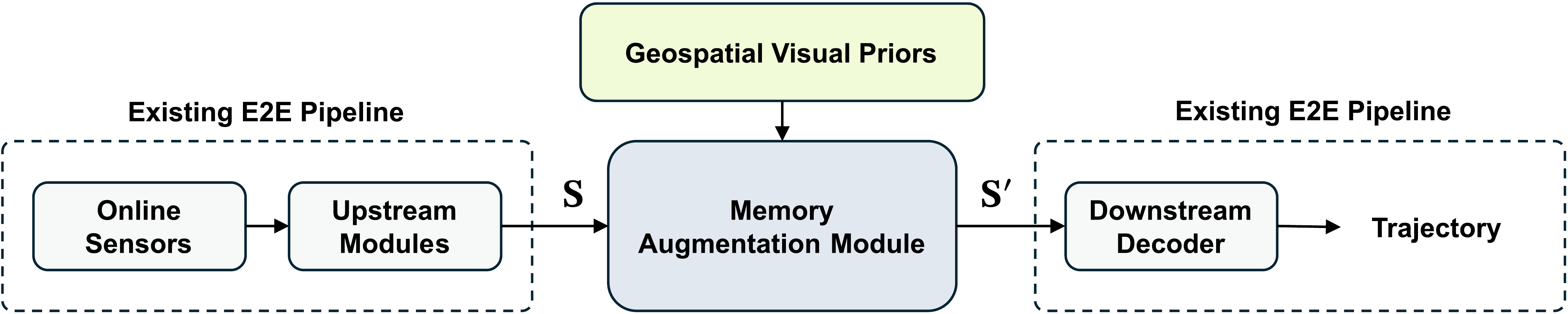}
        \caption{Integration into existing E2E pipelines}
        \label{fig:integration}
    \end{subfigure}
    \caption{\textbf{Overview of the proposed framework.} (a) Detailed architecture of the memory augmentation module. (b) The module takes the current driving state $\mathbf{S}$ from upstream modules and geospatial visual priors as input, and produces an enhanced state $\mathbf{S}'$ for the downstream decoder. Dashed boxes indicate existing components that remain unmodified.}
    \label{fig:architecture_full}
\end{figure}

To incorporate these priors, we introduce the memory augmentation module (Section~\ref{sec:memory_augmentation}; Fig.~\ref{fig:memory_mechanism_detail}), a generic state-editing mechanism that takes the intermediate state of an E2E model along with the retrieved geospatial visual priors and produces an enhanced state. Fig.~\ref{fig:integration} illustrates the general integration scheme into existing E2E pipelines.

\subsection{Geospatial Visual Priors and Memory Bank}
\label{sec:geospatial_visual_priors}

Geospatial visual priors are visual-spatial representations that anchor street-level visual context to the intended driving route, providing the E2E model with foresight of the upcoming environment. To construct these priors, we introduce the memory bank, which stores pairs of embeddings of street-level images and their corresponding locations across the entire map of the target deployment area.

\subsubsection{Construction of the Memory Bank.}
Given a target deployment area, we traverse each lane segment and sample street-level images along the lane centerline at a fixed spatial interval. Each sampled image is encoded by a frozen visual backbone~\cite{tschannenSigLIP2Multilingual2025} and stored as a lane-associated memory. The proposed framework only requires coarsely distributed street-level images and is robust to noise, as the goal is to provide semantically meaningful visual cues for navigation rather than exact geometric reconstruction.

\subsubsection{Querying Geospatial Visual Priors from the Memory Bank.}

During both training and inference, only a subset of visual embeddings is relevant to the current driving context and needs to be retrieved from the memory bank. Given the current vehicle state, we first identify the lane on which the vehicle is currently located. Starting from the current lane, we perform a depth-first search (DFS) over the lane connectivity graph to generate candidate lane sequences extending forward.

Next, based on the high-level navigational intention (\texttt{Turn Left}, \texttt{Turn Right}, and \texttt{Straight}), we select the corresponding lane sequence from the candidate set and retrieve the embeddings stored along it from the memory bank. Each retrieved visual embedding $\mathbf{v}_i \in \mathbb{R}^{D_m}$ is anchored to a relative $(x, y)$ coordinate $\mathbf{x}_i \in \mathbb{R}^{2}$ along the lane, where $D_m$ is the dimensionality of the visual embedding space. Stacking $N$ such pairs yields the visual priors $\mathbf{V} \in \mathbb{R}^{N \times D_m}$ and spatial priors $\mathbf{X} \in \mathbb{R}^{N \times 2}$, which together constitute the queried geospatial visual priors:
\begin{equation}
\{ (\mathbf{v}_i, \mathbf{x}_i) \}_{i=1}^{N}.
\end{equation}

\subsection{Memory Augmentation Module}
\label{sec:memory_augmentation}

We propose a memory augmentation module (Fig.~\ref{fig:memory_mechanism_detail}) to integrate the queried geospatial visual priors into the E2E system. It is composed of two key components: a dual memory architecture that encodes the priors into contextual memory, complemented by a persistent memory, and a memory fusion mechanism that selectively integrates these memories into the current driving state via cross-attention and adaptive gating.

\subsubsection{Dual Memory Architecture.}
\label{sec:dual_memory}

We adopt a dual memory architecture inspired by Titans~\cite{behrouz2025titans}, allowing us to treat priors as retrievable memories. The proposed design decomposes memory into two complementary components: contextual memory, which provides relevant geospatial visual priors, and persistent memory, which serves as a default reference when contextual memory is not helpful.

\paragraph{Contextual Memory.}

Contextual memory is an input-dependent representation that integrates geospatial visual priors. It is constructed as
\begin{equation}
\mathbf{C} = \phi(\mathbf{V}) + \psi(\mathbf{X}),
\end{equation}
where $\phi(\cdot)$ projects the visual priors into the E2E model’s internal embedding space, and $\psi(\cdot)$ denotes a 2D sinusoidal positional embedding that encodes spatial priors in the form of relative coordinates into the same embedding space. Their additive combination enables the contextual memory to jointly capture semantic appearance information and coarse spatial layout.

\paragraph{Persistent Memory.}

However, contextual memory derived from street-level imagery can be corrupted or misaligned due to seasonal changes or geometric mismatches \cite{jiaSpatialRetrievalAugmented2025}. In the absence of alternative references, the attention mechanism is inevitably forced to attend to these corrupted tokens. This issue mirrors the attention sink problem in Transformer architectures \cite{dong2025hymba, xiao2024efficient}, where models require dedicated tokens to absorb attention mass when informative features are missing. Inspired by this, we include persistent memory~\cite{behrouz2025titans, sukhbaatarAugmentingSelfattentionPersistent2019}, a set of learnable tokens designed to serve as a robust fallback against contextual corruption:
\begin{equation}
\mathbf{P} \in \mathbb{R}^{K \times D},
\end{equation}
where $K$ denotes the number of persistent tokens and $D$ is the embedding dimension. These tokens are independent of the current input and shared across all samples.

\subsubsection{Memory Fusion.}
\label{sec:memory_fusion}

To selectively integrate the dual memories into the current driving state, we employ an attention-based mechanism with adaptive gating (Fig.~\ref{fig:memory_mechanism_detail}). We define the current driving state as a set of query tokens
\begin{equation}
\mathbf{S} \in \mathbb{R}^{Q \times D},
\end{equation}
where $Q$ denotes the number of state tokens produced by the E2E driving model and $D$ is the model’s embedding dimension. These tokens encode intermediate driving state features, including perceptual representations and ego-vehicle information.

We model memory retrieval as an associative process using multi-head cross-attention \cite{zhongUnderstandingTransformerPerspective2025, wu2022memorizing,
Sun_2025_CVPR, behrouz2025titans}, with the current driving state as queries and the memory representations as keys and values. Specifically, we concatenate persistent memory and contextual memory along the token dimension and apply cross-attention as follows:
\begin{gather}
\mathbf{M} = [\mathbf{P}; \mathbf{C}], \\
\tilde{\mathbf{S}} = \mathrm{Attn}(\mathbf{S}, \mathbf{M}, \mathbf{M}),
\end{gather}
where $\tilde{\mathbf{S}}$ denotes the memory-attended driving state. This operation allows the model to dynamically attend to both contextual and persistent memory.

\paragraph{Memory Gate.}
We introduce an adaptive gate to regulate memory influence. Drawing from local inference enhancement techniques \cite{chenEnhancedLSTMNatural2017a, conneauSupervisedLearningUniversal2017}, we explicitly model the compatibility between the current state $\mathbf{S}$ and the memory-attended state $\tilde{\mathbf{S}}$ to further improve performance. To measure complementary semantic consistency, we compute the Euclidean distance $d$ \cite{hofferDeepMetricLearning2015} and cosine similarity $c$ \cite{yuan2024rag, wangRADRetrievalAugmentedDecisionMaking2025}:
\begin{gather}
d = \frac{\lVert \mathbf{S} - \tilde{\mathbf{S}} \rVert_2}{\sqrt{D}}, \quad
c = \cos(\mathbf{S}, \tilde{\mathbf{S}}), \\
\mathbf{G} = \sigma\!\left(
f_g\!\left([\mathbf{S}; \tilde{\mathbf{S}}; d; c]\right)
\right),
\end{gather}
where $\mathbf{G}$ denotes the gate, $\sigma(\cdot)$ is the sigmoid function, and $f_g(\cdot)$ is a lightweight MLP. The gate adaptively modulates memory contribution based on the estimated consistency. To stabilize training, we initialize the gate bias toward low values so that it starts in a near-closed state \cite{srivastavaTrainingVeryDeep2015a, gersLearningForgetContinual1999}.

Finally, the fused driving state is obtained via a gated residual update:
\begin{equation}
\mathbf{S}' = \mathbf{S} + \mathbf{G} \odot \tilde{\mathbf{S}},
\end{equation}
where $\odot$ denotes element-wise multiplication. The resulting updated state $\mathbf{S}'$ is then fed back into the downstream decoder of the E2E model, effectively replacing the original state $\mathbf{S}$ with a memory-informed representation for final trajectory prediction.
The proposed module is jointly trained with each baseline model using their original training objectives, without introducing any additional loss terms.

\section{Experiments}
\label{sec:experiments}

\begin{table}[t]
\caption{\textbf{Performance on the NAVSIM-v2 \texttt{navhard-two-stage} benchmark}~\cite{daunerNAVSIMDataDrivenNonReactive2024a, cao2025pseudosimulation}. S1 and S2 denote Stage~1 and Stage~2, respectively. The ``+ PriorEye'' suffix indicates models augmented with our proposed module.}
  \label{tab:performance}
  \centering
  \renewcommand{\arraystretch}{1.2}
  \setlength{\tabcolsep}{3.0pt} 
  
  \resizebox{\textwidth}{!}{
    \begin{NiceTabular}{lcc cccccccc c}[
          code-before = {
            \rowcolor{UltraLightGreen}{4,5,8,9,12,13,16,17} 
          }
        ]
      \toprule

      \textbf{Method} & \textbf{Stg.} & \textbf{NC$\uparrow$} & \textbf{DAC$\uparrow$} & \textbf{DDC$\uparrow$} & \textbf{TLC$\uparrow$} & \textbf{EP$\uparrow$} & \textbf{TTC$\uparrow$} & \textbf{LK$\uparrow$} & \textbf{HC$\uparrow$} & \textbf{EC$\uparrow$} & \textbf{EPDMS$\uparrow$} \\
      \midrule

      \multirow{2}{*}{LTF \cite{chittaTransFuserImitationTransformerBased2023}}
        & S 1 & 95.1 & 80.7 & 98.8 & 99.6 & 84.3 & 94.2 & 92.7 & 97.6 & 79.6 & \multirow{2}{*}{24.7} \\
        & S 2 & 79.4 & 68.3 & 83.4 & 98.1 & 86.6 & 77.0 & 43.4 & 95.9 & 73.2 & \\
      \cmidrule(lr){1-12}
      \multirow{2}{*}{LTF + PriorEye}
        & S 1 & 95.6 & 86.4 & 99.4 & 99.3 & 83.6 & 95.1 & 96.4 & 97.6 & 77.3 & \multirow{2}{*}{\textbf{32.4\,{\scriptsize (+31.2\%)}}} \\
        
        & S 2 & 78.8 & 76.6 & 86.0 & 98.3 & 85.2 & 75.5 & 50.7 & 96.2 & 72.1 & \\
        
      \hline 
      
      \multirow{2}{*}{GTRS-DP \cite{liGeneralizedTrajectoryScoring2025}} 
        & S 1 & 95.0 & 80.7 & 97.0 & 99.9 & 83.1 & 94.1 & 93.1 & 97.6 & 69.9 & \multirow{2}{*}{26.3} \\
        & S 2 & 81.3 & 72.6 & 83.3 & 98.5 & 83.2 & 78.7 & 45.7 & 96.3 & 62.2 & \\
        \cmidrule(lr){1-12}
      \multirow{2}{*}{GTRS-DP + PriorEye} 
        & S 1 &95.3 & 82.0 & 96.4 & 99.8 & 81.5 & 95.1 & 93.1 & 97.1 & 68.0 & \multirow{2}{*}{\textbf{30.1\,{\scriptsize (+14.4\%)}}} \\
        & S 2 & 83.9 & 75.3 & 85.7 & 98.5 & 82.3 & 80.8 & 49.6 & 96.3 & 66.9 & \\
      
      \hline 
      
      \multirow{2}{*}{GTRS-Dense \cite{liGeneralizedTrajectoryScoring2025}} 
        & S 1 & 98.4 & 95.8 & 99.4 & 99.3 & 72.8 & 98.9 & 94.9 & 96.7 & 40.4 & \multirow{2}{*}{44.9} \\
        & S 2 & 91.2 & 89.2 & 94.6 & 98.8 & 68.8 & 90.1 & 55.0 & 93.6 & 49.6 & \\
        \cmidrule(lr){1-12}
      \multirow{2}{*}{GTRS-Dense + PriorEye} 
        & S 1 & 97.6 & 97.1 & 100.0 & 99.8 & 76.1 & 97.8 & 97.3 & 97.8 & 52.4 & \multirow{2}{*}{\textbf{48.6\,{\scriptsize (+8.2\%)}}} \\
        & S 2 & 89.5 & 90.0 & 95.0 & 99.0 & 77.1 & 87.1 & 55.8 & 95.3 & 51.3 & \\
      \hline 
      
      \multirow{2}{*}{DrivoR \cite{kirbyDrivingRegisters2026}} 
        & S 1 & 99.1 & 98.2 & 99.3 & 100.0 & 69.7 & 98.9 & 93.6 & 97.6 & 63.6 & \multirow{2}{*}{48.9} \\
        & S 2 & 91.8 & 89.4 & 94.1 & 99.3 & 59.0 & 89.9 & 51.3 & 98.7 & 76.5 & \\
      \cmidrule(lr){1-12}
      \multirow{2}{*}{DrivoR + PriorEye} 
        & S 1 & 98.7 & 99.6 & 99.6 & 99.8 & 74.5 & 98.4 & 95.6 & 97.6 & 73.8 & \multirow{2}{*}{\textbf{49.6\,{\scriptsize (+1.4\%)}}} \\
        & S 2 & 87.2 & 89.9 & 93.3 & 99.1 & 70.5 & 85.8 & 52.3 & 98.8 & 74.5 & \\

      \bottomrule 
    \end{NiceTabular}
  }
\end{table}

\begin{table}[t]
\caption{\textbf{Performance on the NAVSIM-v2 \texttt{navtest} benchmark}~\cite{daunerNAVSIMDataDrivenNonReactive2024a, cao2025pseudosimulation}.}
  \label{tab:performance_navtest}
  \centering
  \renewcommand{\arraystretch}{1.2}
  \setlength{\tabcolsep}{3.0pt}

  \resizebox{\textwidth}{!}{
    \begin{NiceTabular}{lccccccccc c}[
          code-before = {
            \rowcolor{UltraLightGreen}{3,5,7,9}
          }
        ]
      \toprule 
      \textbf{Method} & \textbf{NC$\uparrow$} & \textbf{DAC$\uparrow$} & \textbf{DDC$\uparrow$} & \textbf{TLC$\uparrow$} & \textbf{EP$\uparrow$} & \textbf{TTC$\uparrow$} & \textbf{LK$\uparrow$} & \textbf{HC$\uparrow$} & \textbf{EC$\uparrow$} & \textbf{EPDMS$\uparrow$} \\
      \midrule 

      LTF \cite{chittaTransFuserImitationTransformerBased2023}
        & 97.8 & 92.5 & 99.1 & 99.8 & 87.8 & 97.4 & 96.3 & 98.3 & 87.0 & 84.3 \\
      LTF + PriorEye
        & 97.6 & 95.1 & 99.5 & 99.8 & 87.8 & 97.4 & 97.4 & 98.3 & 87.0 & \textbf{86.8\,{\scriptsize (+2.5)}} \\
      \cmidrule(lr){1-11}

      GTRS-DP \cite{liGeneralizedTrajectoryScoring2025}
        & 97.3 & 92.3 & 98.6 & 99.7 & 86.4 & 97.0 & 95.3 & 98.1 & 79.4 & 82.2 \\
      GTRS-DP + PriorEye
        & 97.5 & 92.7 & 98.7 & 99.8 & 85.9 & 97.1 & 95.6 & 98.2 & 78.3 & \textbf{82.5\,{\scriptsize (+0.3)}} \\
      \cmidrule(lr){1-11}

      GTRS-Dense \cite{liGeneralizedTrajectoryScoring2025}
        & 99.1 & 98.3 & 99.6 & 99.9 & 82.8 & 99.2 & 94.8 & 98.0 & 47.5 & 85.4 \\
      GTRS-Dense + PriorEye
        & 98.7 & 99.1 & 99.8 & 99.9 & 85.0 & 98.8 & 96.7 & 98.2 & 67.5 & \textbf{88.8\,{\scriptsize (+3.4)}} \\
      \cmidrule(lr){1-11}

      DrivoR \cite{kirbyDrivingRegisters2026}
        & 99.4 & 99.2 & 99.8 & 99.9 & 76.6 & 99.4 & 94.9 & 98.3 & 71.7 & 87.2 \\
      DrivoR + PriorEye
        & 99.1 & 99.6 & 99.8 & 99.9 & 81.5 & 99.1 & 95.9 & 98.3 & 82.5 & \textbf{89.9\,{\scriptsize (+2.7)}} \\

      \bottomrule 
    \end{NiceTabular}
  }
\end{table}

\subsection{Experimental Setup}

\subsubsection{Dataset.}
We train and evaluate our method on the NAVSIM v2 benchmark~\cite{daunerNAVSIMDataDrivenNonReactive2024a, cao2025pseudosimulation}, a large-scale real-world end-to-end autonomous driving benchmark built upon nuPlan \cite{karnchanachariLearningbasedPlanningNuPlan2024}. All models are trained on the full \texttt{navtrain} split and evaluated on the \texttt{navhard-two-stage} and \texttt{navtest} splits, following the official benchmark protocol. For ablation studies, we reserve a subset of \texttt{navtrain} for validation, training models on the remaining data and evaluating them on the held-out validation split.

\subsubsection{Metrics.}
We report performance using the Extended Predictive Driver Model Score (EPDMS), the official composite metric of the NAVSIM v2 benchmark~\cite{daunerNAVSIMDataDrivenNonReactive2024a, cao2025pseudosimulation}. EPDMS aggregates several safety and comfort sub-metrics: No-at-fault Collision (NC), Drivable Area Compliance (DAC), Driving Direction Compliance (DDC), Traffic Light Compliance (TLC), Ego Progress (EP), Time-to-Collision (TTC), Lane Keeping (LK), History Comfort (HC), and Extended Comfort (EC). The score for \texttt{navhard-\allowbreak two-\allowbreak stage} is computed via a two-stage pseudo-simulation protocol: Stage~1 evaluates planned trajectories on real-world observations, while Stage~2 assesses robustness under controlled perturbations using synthetic observations. In contrast, \texttt{navtest} follows only the Stage~1 evaluation.

\subsubsection{Baselines and Terminology.}
We evaluate the model-agnostic nature of our proposed framework by integrating it into four representative E2E driving
baselines spanning diverse paradigms: regression-based (LTF~\cite{chittaTransFuserImitationTransformerBased2023}), diffusion-based 
(GTRS-DP~\cite{liGeneralizedTrajectoryScoring2025}), and scoring-based (GTRS-Dense~\cite{liGeneralizedTrajectoryScoring2025} and 
DrivoR~\cite{kirbyDrivingRegisters2026}). Models augmented with our proposed module are denoted with the ``+ PriorEye'' suffix. For qualitative results and robustness evaluations, we default to GTRS-Dense as the baseline and GTRS-Dense + PriorEye as our method. For ablation studies, we use LTF as the baseline, as its lightweight training enables efficient exploration of the many design configurations.

\subsubsection{Street-Level Imagery.}
We leverage the Google Street View (GSV) Image API to query street-level imagery for the covered regions, which serves as the source for constructing the memory bank. GSV imagery is not available at every location and does not guarantee precise spatial alignment. As a result, some queried locations may return no image, while others may suffer from vertical (z-axis) misalignment or spatial mislocalization, causing multiple distinct positions along a lane to be mapped to the same street-view image~\cite{jiaSpatialRetrievalAugmented2025}. In our setting, such missing or redundant observations are not problematic, as discussed in Section~\ref{sec:dual_memory}. Although we choose GSV as it provides a convenient and scalable source of street-level imagery that covers the regions associated with our dataset, the proposed framework is not tied to a specific API or data provider. Any sparsely sampled collection of location-anchored driving images can serve the same role.

\subsubsection{Implementation Details.}
For the memory bank, we encode street-level imagery using SigLIP2~\cite{tschannenSigLIP2Multilingual2025} (base-patch16-256) at a fixed spatial interval of 5 meters, resulting in a total size of 939~MB covering approximately 6.5~km$^2$ of the target area~\cite{karnchanachariLearningbasedPlanningNuPlan2024}.

At inference, $N=20$ geospatial visual priors are retrieved, corresponding to a look-ahead range of approximately 100 meters. The memory augmentation module introduces only 713K additional parameters, representing a negligible overhead of 1.3\%, 0.6\%, 0.9\%, and 1.7\% relative to LTF (56.7M), GTRS-DP (116M), GTRS-Dense (83.6M), and DrivoR (41.5M), respectively.

For each baseline, the memory augmentation module edits the intermediate state formed by concatenating the encoded ego-status with perceptual features specific to each architecture: BEV features for LTF and GTRS-DP, image features for GTRS-Dense, and scene tokens for DrivoR.

For training, we follow the existing protocols of each baseline~\cite{chittaTransFuserImitationTransformerBased2023, liGeneralizedTrajectoryScoring2025, kirbyDrivingRegisters2026}. All models are trained on 8 NVIDIA RTX 5090 GPUs with per-GPU batch sizes of 64, 16, 16, and 12 for LTF, GTRS-DP, GTRS-Dense, and DrivoR, trained for 100, 80, 40, and 20 epochs, respectively.

\subsection{Main Results}

\begin{figure}[!t]
  \centering
  \includegraphics[width=\textwidth]{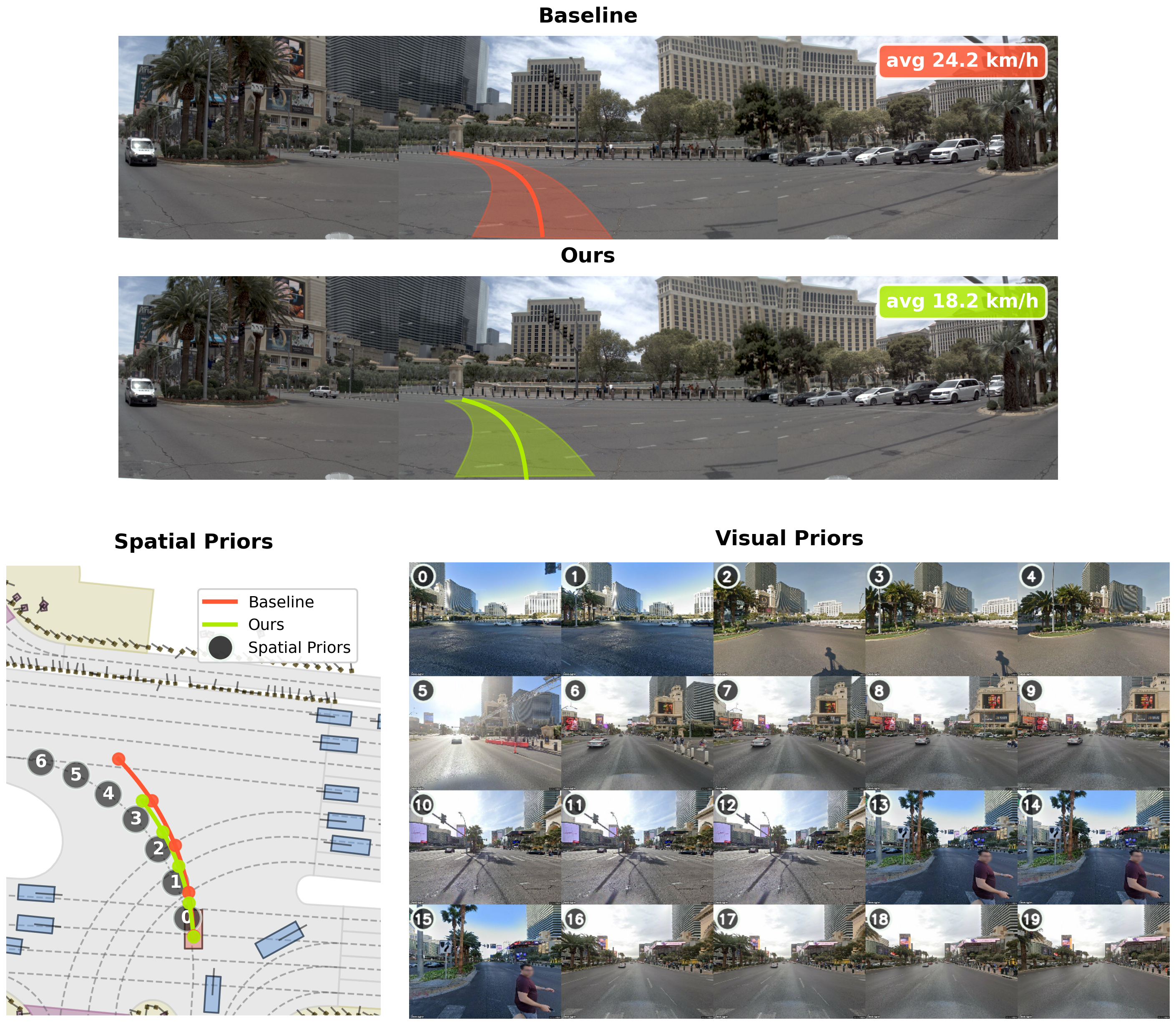}
  \caption{\textbf{Qualitative visualization of planned trajectories in a left-turn scenario.}
  \textit{Top:} Projected future trajectories overlaid on the camera view. The average speed over the prediction horizon is shown in the top-right corner.
  \textit{Bottom-Left:} BEV map showing the predicted trajectories and spatial priors sampled at 5-meter intervals.
  \textit{Bottom-Right:} Retrieved street-level visual priors indexed along the intended route.
  In this scenario, the downstream road contains a pedestrian crosswalk not yet visible to onboard sensors. Relying solely on instantaneous observations, the baseline (red) maintains an average speed of 24.2~km/h. In contrast, our method (green) leverages geospatial visual priors to anticipate the upcoming crosswalk, reflected in the retrieved memories (indices 10--15), and drives at a more cautious 18.2~km/h. This demonstrates how geospatial priors effectively complement instantaneous perception to enable anticipatory driving behavior.}
  \label{fig:visualization_results}
\end{figure}

Table~\ref{tab:performance} reports the results on the NAVSIM-v2 \texttt{navhard-\allowbreak two-\allowbreak stage} benchmark. Across all four end-to-end driving baselines, incorporating the proposed module consistently improves performance, demonstrating the model-agnostic effectiveness of our approach. Specifically, memory augmentation yields relative EPDMS improvements of 31.2\%, 14.4\%, 8.2\%, and 1.4\% over LTF, GTRS-DP, GTRS-Dense, and DrivoR, respectively.

We observe a consistent improvement on the \texttt{navtest} split (Table~\ref{tab:performance_navtest}), where PriorEye again improves all four baselines, with EPDMS gains of 2.5, 0.3, 3.4, and 2.7 points. This consistency across both splits confirms that the benefit of geospatial visual priors is not specific to a single evaluation setting.

Beyond quantitative gains, geospatial visual priors effectively complement instantaneous perception under partial observability, such as occluded intersections. A qualitative example is provided in Fig.~\ref{fig:visualization_results}. By anticipating upcoming road features before they become visible, the model can maintain a more conservative speed, allowing for smoother deceleration when potential hazards arise.

\subsection{Robustness Analysis}
We further evaluate robustness under (i) instantaneous sensor corruption and (ii) corrupted geospatial visual priors.

\subsubsection{Sensor Robustness.}
\label{sec:sensor_robustness}

\begin{figure}
    \centering
    \includegraphics[width=0.32\linewidth]{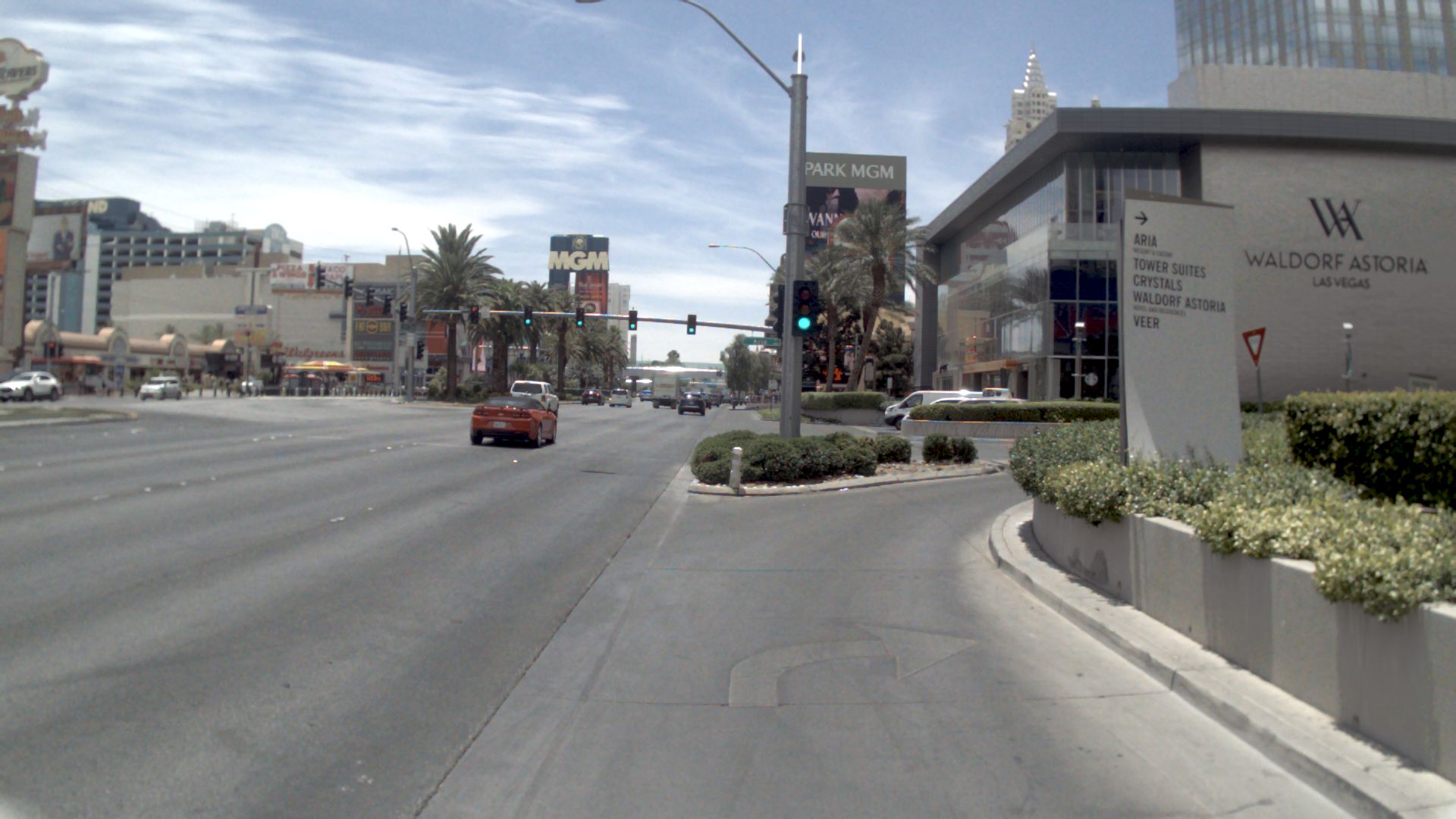}
    \includegraphics[width=0.32\linewidth]{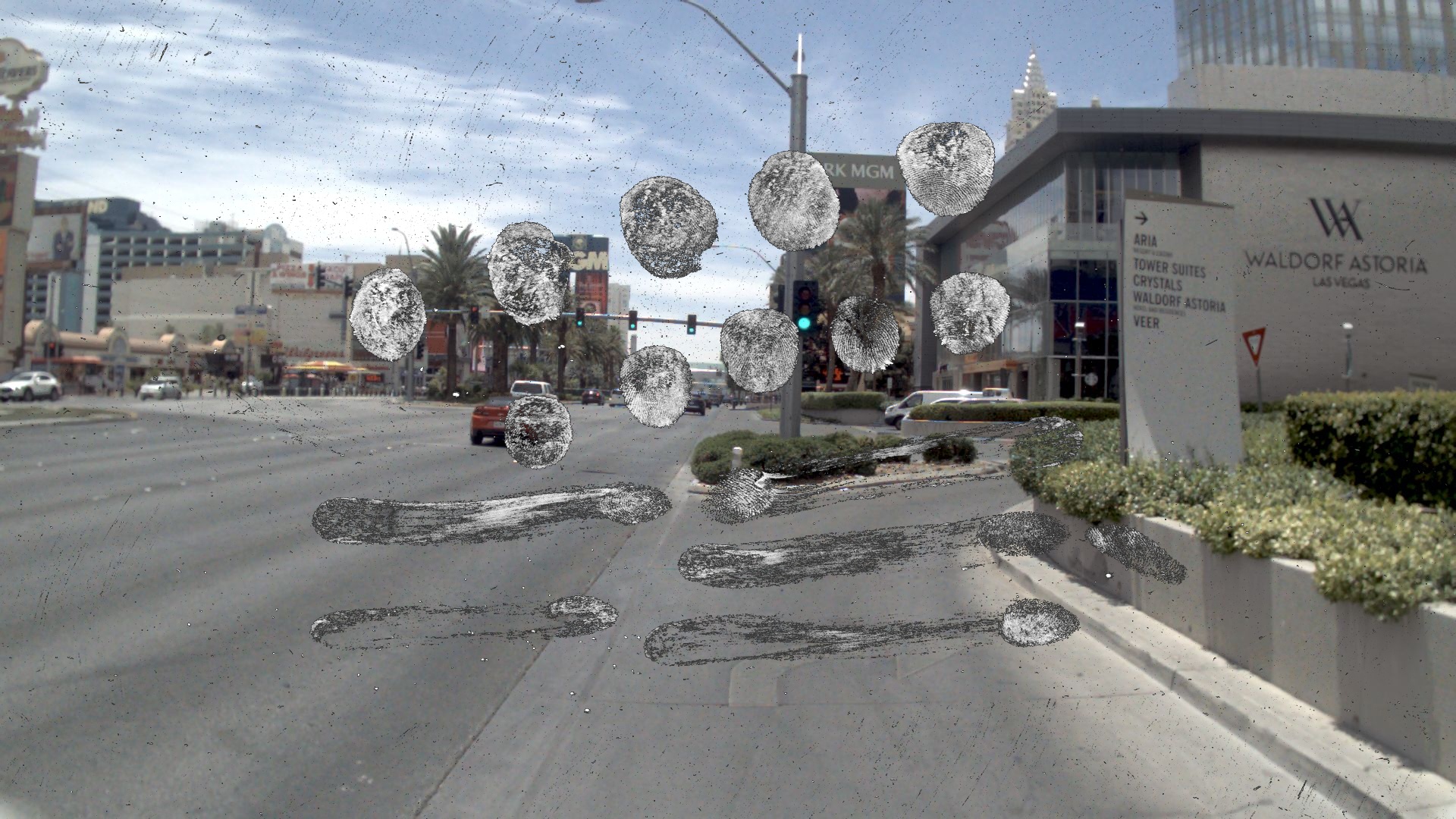}
    \includegraphics[width=0.32\linewidth]{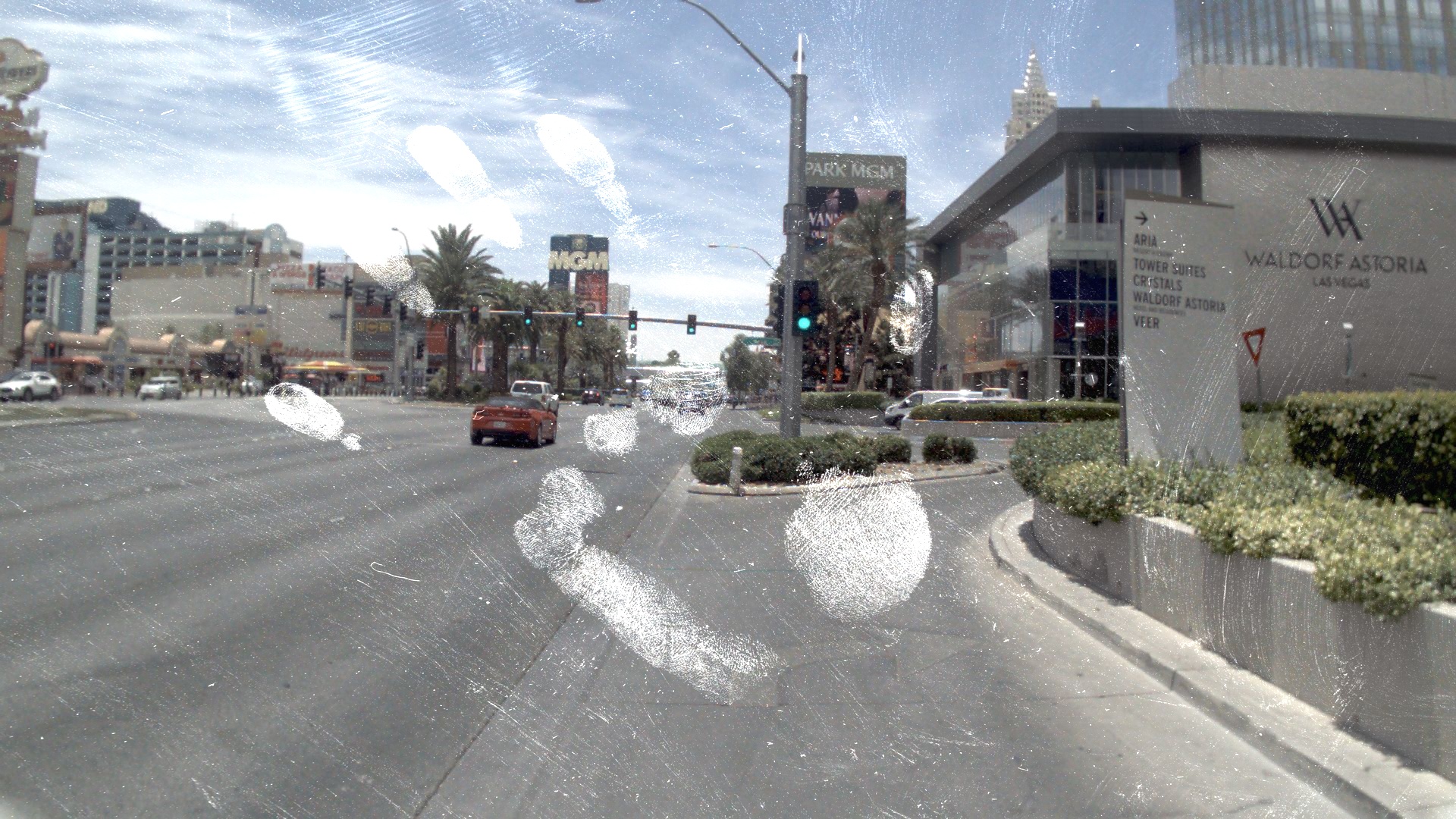}
    \\[4pt]
    \includegraphics[width=0.32\linewidth]{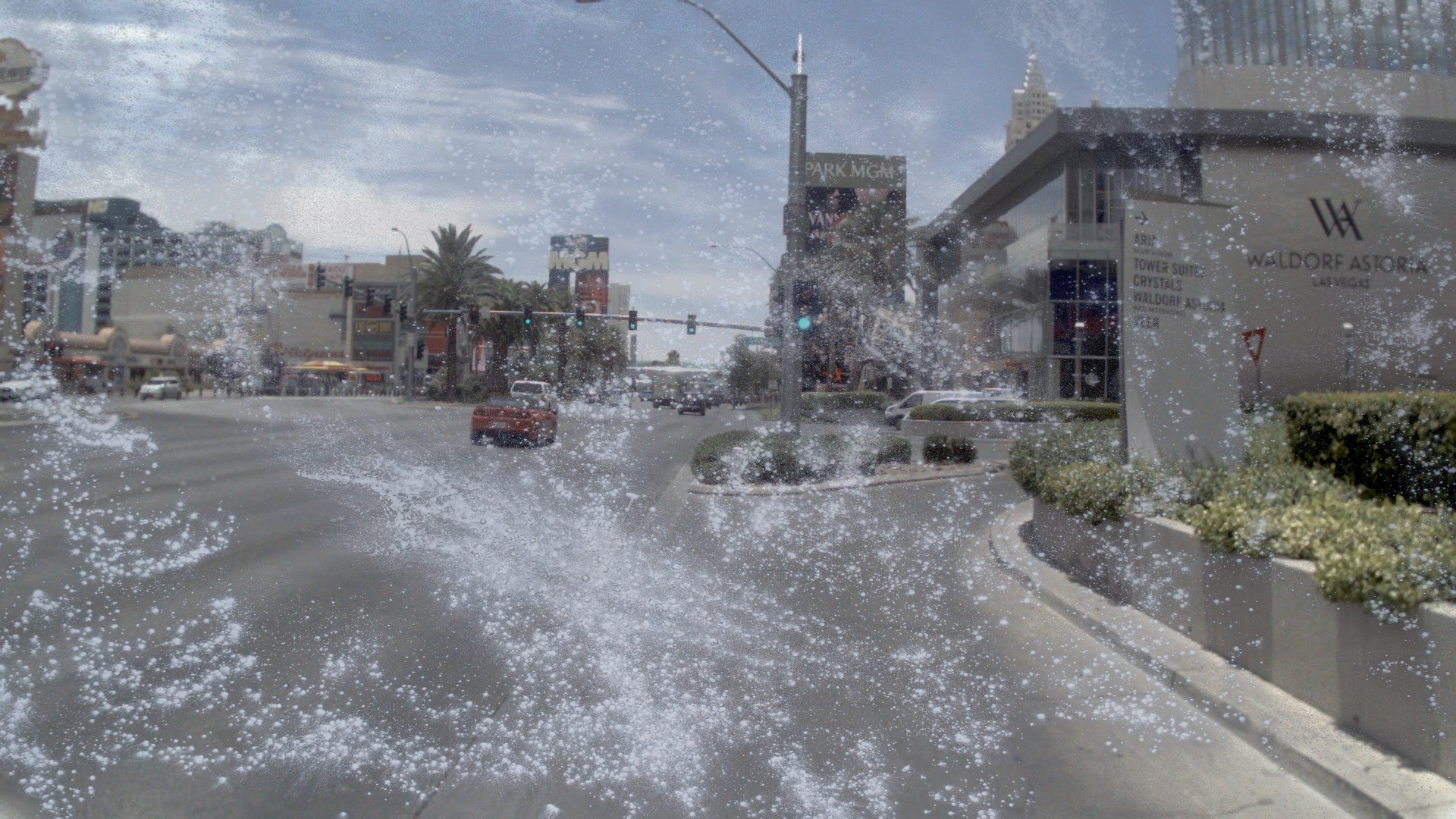}
    \includegraphics[width=0.32\linewidth]{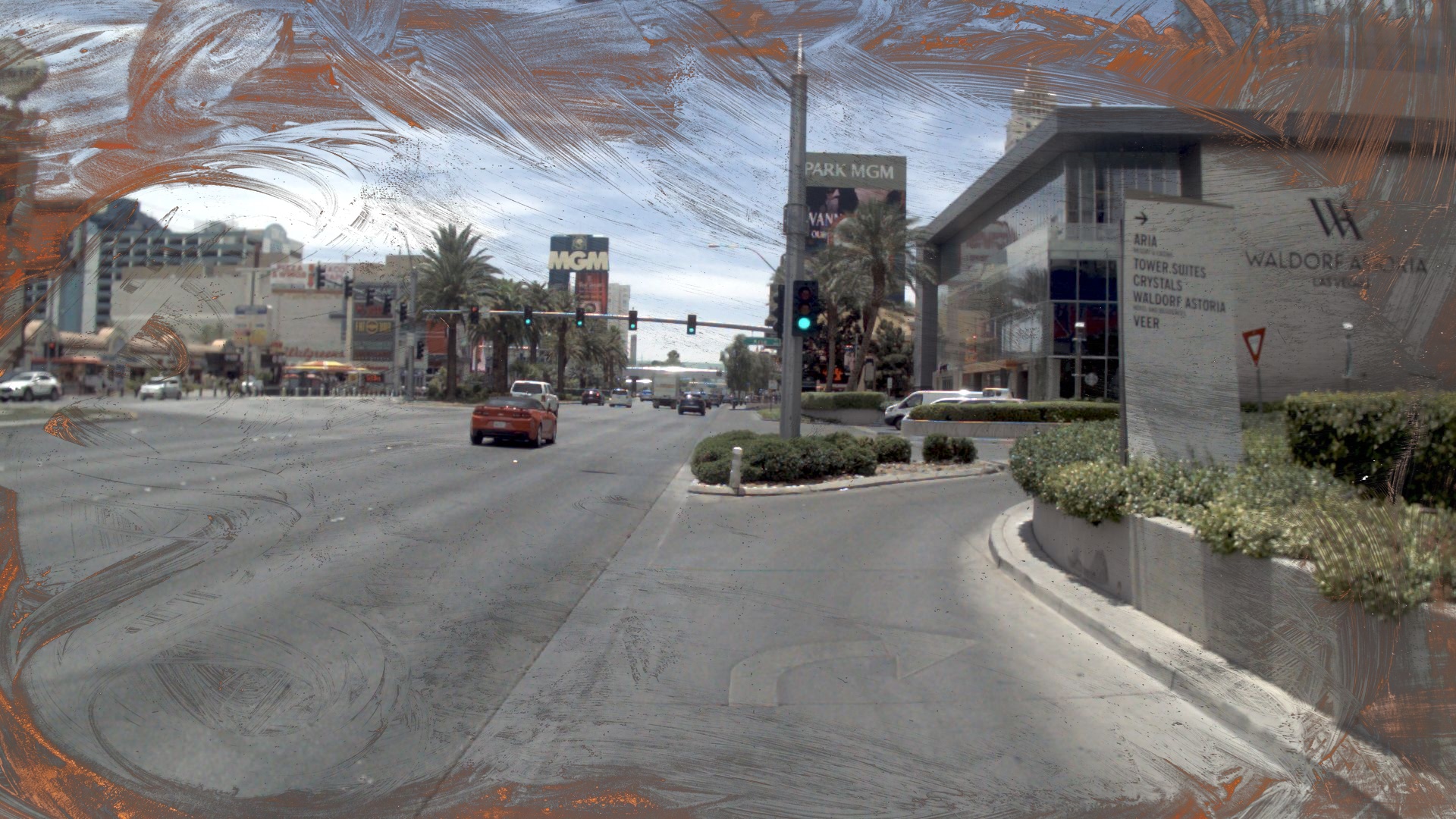}
    \includegraphics[width=0.32\linewidth]{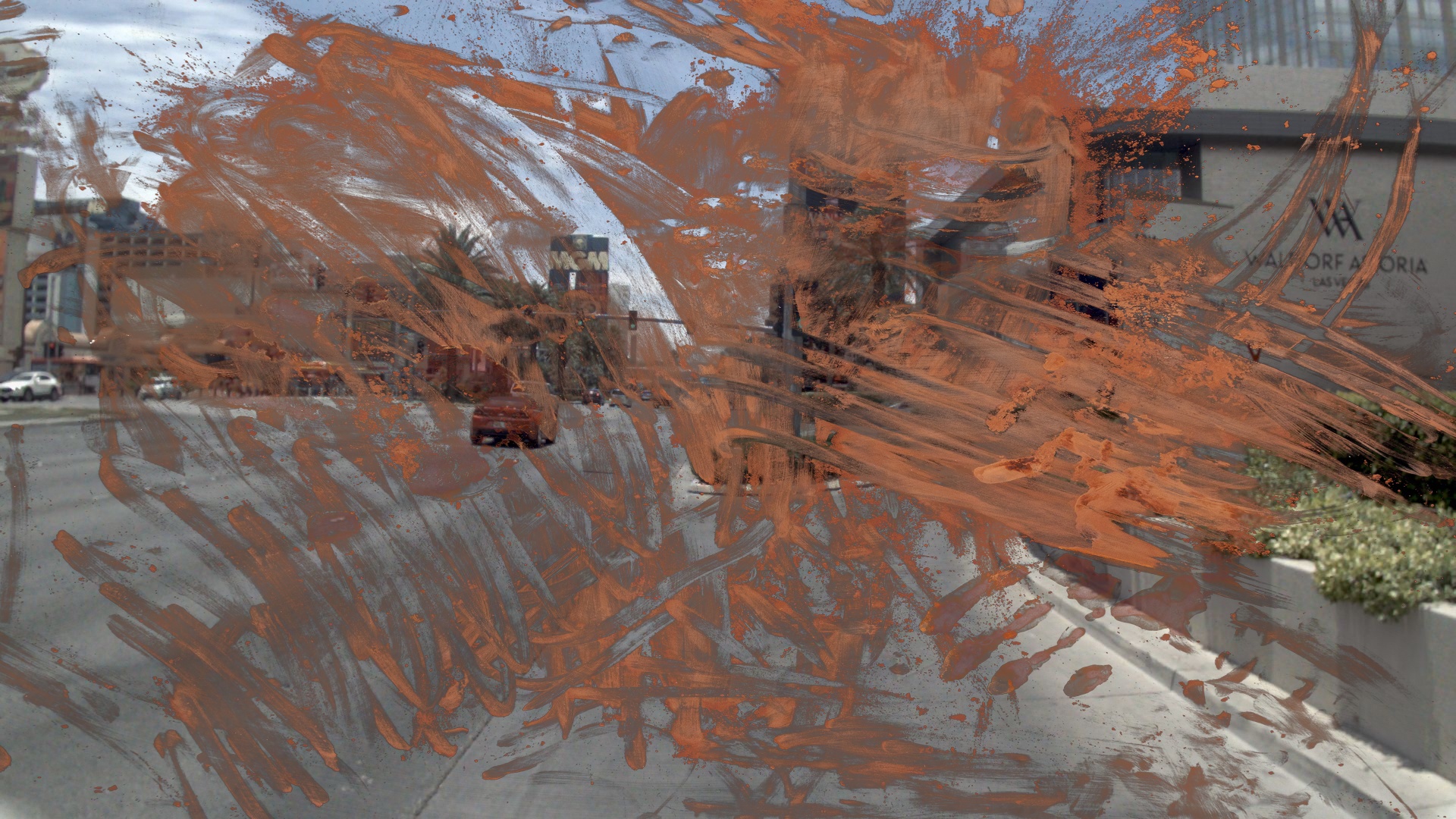}
    \caption{Example of synthetic sensor corruptions applied to the front-center camera. From left to right, top to bottom: Nominal, Fingerprints, Handprints, Frost, Mud (Mild), Mud (Heavy).}
    \label{fig:sensor_corruptions}
\end{figure}

Sensor corruption is a common and unavoidable challenge in real-world autonomous driving~\cite{ceccarelliRGBCamerasFailures2023}. This is particularly challenging for current E2E systems relying on instantaneous data, as the only source of information about the environment then becomes unreliable. We hypothesize that geospatial visual priors can improve robustness in such conditions, as retrieved driving memory remains unaffected by onboard sensor degradation. To verify this, we synthetically apply five different corruptions to camera observations. This is achieved by using off-the-shelf transparent overlays captured from a real camera and representing different sensor degradations. Each overlay is combined with the original images by using additive blending (Linear Dodge) for handprints, or alpha compositing for all other corruptions. Figure~\ref{fig:sensor_corruptions} shows examples of each synthetic corruption applied to the front-center camera.

Table~\ref{tab:sensor_robustness} reports the quantitative robustness results. Our method consistently outperforms the baseline across all corruption types. The improvement is most pronounced under severe mud corruption, where the baseline suffers a 53.2\% performance drop while our method reduces this to 32.5\%. Even under mild corruptions such as fingerprints and handprints, our method achieves notably smaller degradation. On average, our method reduces the relative performance degradation from 20.6\% to 13.9\% across all corruption types. These results demonstrate that geospatial visual priors provide a complementary source of information, improving robustness across diverse corruption scenarios.

\begin{table}
    \centering
    \caption{\textbf{Sensor robustness evaluation on \texttt{navhard-\allowbreak two-\allowbreak stage} (EPDMS).} Degradation relative to nominal is shown in parentheses.}
    \label{tab:sensor_robustness}
    \renewcommand{\arraystretch}{1.1}
    \setlength{\tabcolsep}{6pt}
    \resizebox{\linewidth}{!}{
        \begin{NiceTabular}{lcccccc}[
            code-before = {
                \rowcolor{UltraLightGreen}{3}
            }
        ]
            \toprule
             & \textbf{Nominal} & \textbf{Fingerprints} & \textbf{Handprints} & \textbf{Frost} & \textbf{Mud (Mild)} & \textbf{Mud (Heavy)} \\
            \midrule
            Baseline & 44.9 & 40.0 {\scriptsize (-10.9\%)} & 40.3 {\scriptsize (-10.2\%)} & 34.6 {\scriptsize (-22.9\%)} & 42.3 {\scriptsize (-5.8\%)} & 21.0 {\scriptsize (-53.2\%)} \\
            Ours & \textbf{48.6} & \textbf{45.1 {\scriptsize (-7.2\%)}} & \textbf{45.4 {\scriptsize (-6.6\%)}} & \textbf{38.1 {\scriptsize (-21.6\%)}} & \textbf{47.9 {\scriptsize (-1.4\%)}} & \textbf{32.8 {\scriptsize (-32.5\%)}} \\
            \bottomrule
        \end{NiceTabular}
    }
\end{table}

\subsubsection{Geospatial Visual Prior Corruption.}
In real-world deployment, geospatial visual priors may become unreliable due to outdated imagery, appearance drift, spatial misalignment, or missing observations. Since contextual memory ($\mathbf{C}$) is constructed from retrieved visual--spatial pairs $(\mathbf{V}, \mathbf{X})$, such failures correspond to corruption of $\mathbf{C}$. As discussed in Section~\ref{sec:dual_memory}, our framework incorporates persistent memory ($\mathbf{P}$) to mitigate this. To evaluate this design choice, we simulate three corruption scenarios. For visual corruption, we replace the retrieved visual embeddings with those queried from a randomly sampled location 500 meters away, introducing semantically irrelevant appearance features. For spatial corruption, we replace the associated spatial coordinates with random positions at the same 500-meter offset, while keeping the visual embeddings intact. The combined corruption applies both perturbations simultaneously.

Table~\ref{tab:prior_corruption} reports the attention behavior of the full dual-memory model and the driving performance across ablation variant models under corruption. The model adaptively reallocates attention towards the fallback persistent memory $\mathbf{P}$ as contextual memory becomes unreliable, reaching 0.97 under visual corruption and 0.99 under combined corruption, effectively suppressing misleading contextual cues.

Under normal conditions, both the context-only model and the full dual-memory model outperform the baseline. Under corruption, however, the context-only model suffers severe degradation, whereas the dual-memory model remains robust, reducing performance drops from 19.6\% to 6.8\% under visual corruption and from 18.6\% to 6.6\% under combined corruption. Notably, even under corrupted priors, our full model consistently outperforms the baseline. These results highlight the critical role of persistent memory $\mathbf{P}$ as a reliable fallback when contextual information becomes unreliable, demonstrating that the proposed dual-memory framework is inherently robust to geospatial visual prior corruption.

\begin{table}
\footnotesize
\centering
\caption{\textbf{Robustness under corrupted geospatial visual priors on \texttt{navhard-\allowbreak two-\allowbreak stage}.} Left: average attention weights measured in the full dual-memory model ($\mathbf{P}+\mathbf{C}$). Right: EPDMS for the baseline, the model with contextual memory only ($\mathbf{C}$ only), and the full model ($\mathbf{P}+\mathbf{C}$). Degradation relative to the uncorrupted condition is shown in parentheses.}
\label{tab:prior_corruption}
\renewcommand{\arraystretch}{1.0}
\setlength{\tabcolsep}{5pt}
\begin{NiceTabular}{l cc | ccc}
\toprule
& \multicolumn{2}{c}{\textbf{Attention}} & \multicolumn{3}{c}{\textbf{EPDMS}} \\
\cmidrule(lr){2-3} \cmidrule(lr){4-6}
\textbf{Corruption} & \textbf{to $\mathbf{P}$} & \textbf{to $\mathbf{C}$} & \textbf{Baseline} & \textbf{$\mathbf{C}$ only} & \textbf{$\mathbf{P}+\mathbf{C}$} \\
\midrule
None    & 0.29 & \textbf{0.71} & 44.9 & 47.9 & \textbf{48.6} \\
Visual  & \textbf{0.97} & 0.03 & -- & 38.5 {\scriptsize (-19.6\%)} & \textbf{45.3 {\scriptsize (-6.8\%)}} \\
Spatial & \textbf{0.65} & 0.35 & -- & 45.2 {\scriptsize (-5.6\%)} & \textbf{46.6 {\scriptsize (-4.1\%)}} \\
Both    & \textbf{0.99} & 0.01 & -- & 39.0 {\scriptsize (-18.6\%)} & \textbf{45.4 {\scriptsize (-6.6\%)}} \\
\bottomrule
\end{NiceTabular}
\end{table}

\subsection{Ablation Studies}
\label{sec:ablation}

\begin{table}[ht]
   \footnotesize
  \centering
    \caption{\textbf{Ablation studies on the validation split.} For reference, baseline (LTF) scores 81.7 EPDMS.}
  \label{tab:ablation}
  \renewcommand{\arraystretch}{1.0}
  \setlength{\tabcolsep}{6pt} 
  
  \begin{NiceTabular}{l l c}
    \toprule
    \textbf{Description} & \textbf{Configuration} & \textbf{EPDMS} $\uparrow$ \\
    \midrule
    \multirow{3}{*}{Embedding backbone}
      & DINOv2 & 83.4 \\
      & SegFormer & 82.9 \\
      & \textbf{SigLIP2} & \textbf{84.4} \\
    \midrule
    \multirow{2}{*}{Retrieval strategy}
      & Proximity-based & 82.1 \\
      & \textbf{Intention-guided} & \textbf{84.4} \\
    \midrule
    \multirow{3}{*}{Memory components}
      & $\mathbf{P}$ only & 82.2 \\
      & $\mathbf{C}$ only & 83.9 \\
      & \textbf{$\mathbf{P}+\mathbf{C}$} & \textbf{84.4} \\
    \midrule
    \multirow{3}{*}{Contextual memory components}
      & $\mathbf{C} = \psi(\mathbf{X})$ & 82.7 \\
      & $\mathbf{C} = \phi(\mathbf{V})$ & 84.0 \\
      & \textbf{$\mathbf{C} = \phi(\mathbf{V}) + \psi(\mathbf{X})$} & \textbf{84.4} \\
    \midrule
    \multirow{4}{*}{Memory gate components}
      & $\mathbf{G} = \sigma\!\left(f_g\!\left([\mathbf{S}; \tilde{\mathbf{S}}]\right)\right)$ & 83.0 \\
      & $\mathbf{G} = \sigma\!\left(f_g\!\left([\mathbf{S}; \tilde{\mathbf{S}}; d]\right)\right)$ & 83.4 \\
      & $\mathbf{G} = \sigma\!\left(f_g\!\left([\mathbf{S}; \tilde{\mathbf{S}}; c]\right)\right)$ & 83.4 \\
      & \textbf{$\mathbf{G} = \sigma\!\left(f_g\!\left([\mathbf{S}; \tilde{\mathbf{S}}; d; c]\right)\right)$} & \textbf{84.4} \\
  \bottomrule
  \end{NiceTabular}%
\end{table}

Finally, we validate all design details of our proposed approach through extensive ablation studies, as summarized in Table~\ref{tab:ablation}. All ablations use LTF as the baseline and are evaluated on the validation split.

\paragraph{Embedding backbone.}
All evaluated backbones for encoding geospatial visual priors improve over the baseline, confirming the value of priors regardless of the encoder. Among the evaluated models, SigLIP2~\cite{tschannenSigLIP2Multilingual2025} outperforms both DINOv2~\cite{oquab2024dinov} and SegFormer~\cite{xie2021segformer}, suggesting that its richer semantic representations are better suited for encoding street-level visual priors.

\paragraph{Retrieval strategy.}

The retrieval strategy for geospatial visual priors plays a crucial role. A naive proximity-based strategy~\cite{jiaSpatialRetrievalAugmented2025} that retrieves the $N$ nearest prior nodes yields only a marginal gain over the baseline, suggesting that spatially close but maneuver-irrelevant priors provide little useful signal. In contrast, our intention-guided retrieval selects priors conditioned on the planned maneuver, yielding a substantially larger improvement.

\paragraph{Memory components.}
While contextual memory $\mathbf{C}$ alone already provides a clear gain over the baseline, persistent memory $\mathbf{P}$ alone yields only a marginal improvement, suggesting that adding learnable parameters alone is not sufficient without meaningful contextual input. Combining both yields the best performance, supporting the dual-memory formulation that benefits not only robustness, as discussed in Section~\ref{sec:sensor_robustness}, but also nominal performance.

\paragraph{Contextual memory components.}
Spatial encodings $\psi(\mathbf{X})$ provide geometric guidance via the centerline of the intention-guided target lane, yielding a modest gain of 1.0 EPDMS over the baseline. Visual encodings $\phi(\mathbf{V})$ contribute more substantially with a gain of 2.3 EPDMS, offering rich appearance details of the upcoming path. Combining both yields the largest gain of 2.7 EPDMS, demonstrating that structural geometry and semantic appearance are complementary.

\paragraph{Memory gate components.}
Incorporating similarity cues, namely cosine similarity $c$ and normalized Euclidean distance $d$, individually improves EPDMS, and using both cues together achieves the best performance. This confirms that the two metrics capture complementary aspects of compatibility between the current state and the retrieved memory.

\subsection{Runtime Analysis}
\label{sec:runtime}
We analyze the inference overhead introduced by PriorEye on top of the GTRS-Dense baseline as summarized in \cref{tab:runtime}. All measurements use a single sample (batch size 1) on an NVIDIA RTX PRO 6000, with CPU-side feature computation on an AMD Ryzen Threadripper 7960X. The memory augmentation module adds only $5.6\,\mathrm{ms}$ of GPU latency, and the CPU-side memory retrieval adds $3.7\,\mathrm{ms}$ to feature computation. The total inference time remains $67.2\,\mathrm{ms}$, comfortably within a $10\,\mathrm{Hz}$ runtime budget. Since the SigLIP2 encoding of street-level imagery is performed entirely offline during memory-bank construction, it incurs no overhead at inference time.

\begin{table}[ht]
  \centering
  \footnotesize
  \caption{\textbf{Inference runtime of PriorEye on GTRS-Dense.} Values are averaged over 495 GPU inference runs (after 5 warm-up runs) and 200 CPU feature-computation runs.}
  \label{tab:runtime}
  \begin{NiceTabular}{l c c c}
    \toprule
    & \textbf{CPU feat. (ms)} & \textbf{GPU (ms)} & \textbf{Total (ms)} \\
    \midrule
    GTRS-Dense          & 15.5 & 42.4 & 57.9 \\
    \,+ PriorEye        & 19.2 & 48.0 & 67.2 \\
    \midrule
    Overhead            & +3.7 & +5.6 & +9.3 \\
    \bottomrule
  \end{NiceTabular}
\end{table}

\section{Conclusion}
\label{sec:conclusion}
We introduced geospatial visual priors to equip E2E driving policies with visual-spatial foresight beyond the immediate sensor view. Our model-agnostic memory augmentation module, featuring dual memory and adaptive gating, consistently improves performance on the NAVSIM-v2 benchmark. In addition, the proposed method provides robustness under sensor degradation, while ensuring resilience when retrieved priors become unreliable.

\subsubsection{Limitations.} Our current design has two main limitations that suggest promising future directions. First, prior retrieval relies solely on ego-vehicle location. Incorporating appearance-based matching could enable more robust retrieval, allowing the system to recall routes using both spatial and visual cues. Second, the memory bank is constructed from static street-level imagery. With repeated driving logs over the same area, the framework could be extended with memory update and forgetting mechanisms, enabling the system to continuously adapt to long-term environmental changes.

\section*{Acknowledgements}
The work was supported by the EPSRC Programme Grant ``From Sensing to Collaboration'' (EP/V000748/1).

%
%
\bibliographystyle{splncs04}
\bibliography{short_ref}

\newpage
\appendix

\begingroup
\centering
\section*{\centering PriorEye: Geospatial Visual Priors for End-to-End Autonomous Driving}
\large Supplementary Material\\[1em]
\endgroup

\section{EPDMS Metric Details}
\label{sec:supp_epdms}

We adopt the Extended Predictive Driver Model Score (EPDMS) from NAVSIM v2~\cite{daunerNAVSIMDataDrivenNonReactive2024a, cao2025pseudosimulation}. EPDMS consists of four multiplier metrics and five weighted metrics, summarized in Table~\ref{tab:epdms_metrics}. The score is computed as:
\begin{equation}
  \text{EPDMS} = \prod_{m \in \mathcal{M}_{\text{mult}}} \tilde{m} \;\cdot\; \frac{\sum_{m \in \mathcal{M}_{\text{wt}}} w_m \cdot \tilde{m}}{\sum_{m \in \mathcal{M}_{\text{wt}}} w_m},
  \label{eq:epdms}
\end{equation}
where $m_{\text{agent}}$ and $m_{\text{human}}$ denote the metric score of the planner and the human driver for a given scenario, respectively, and $\tilde{m}$ denotes a false-positive filtered score that exempts the planner when the human driver also commits the same violation:
\begin{equation}
  \tilde{m} = 
  \begin{cases}
    1.0 & \text{if } m_{\text{human}} = 0, \\
    m_{\text{agent}} & \text{otherwise}.
  \end{cases}
  \label{eq:filter}
\end{equation}
The final benchmark uses a two-stage evaluation: Stage~1 evaluates the initial scene over a 4-second horizon, and Stage~2 evaluates follow-up scenes to approximate closed-loop behavior. The two stages are multiplied to produce the final score.

\begin{table}[h]
  \caption{\textbf{EPDMS metric composition}~\cite{cao2025pseudosimulation}.}
  \label{tab:epdms_metrics}
  \centering
  \begin{tabular}{llcc}
    \toprule
    Metric & Abbr. & Type & Range \\
    \midrule
    No at-fault Collisions & NC & multiplier & $\{0, \tfrac{1}{2}, 1\}$ \\
    Drivable Area Compliance & DAC & multiplier & $\{0, 1\}$ \\
    Driving Direction Compliance & DDC & multiplier & $\{0, \tfrac{1}{2}, 1\}$ \\
    Traffic Light Compliance & TLC & multiplier & $\{0, 1\}$ \\
    \midrule
    Ego Progress & EP & weighted ($w{=}5$) & $[0, 1]$ \\
    Time to Collision & TTC & weighted ($w{=}5$) & $\{0, 1\}$ \\
    Lane Keeping & LK & weighted ($w{=}2$) & $\{0, 1\}$ \\
    History Comfort & HC & weighted ($w{=}2$) & $\{0, 1\}$ \\
    Extended Comfort & EC & weighted ($w{=}2$) & $\{0, 1\}$ \\
    \bottomrule
  \end{tabular}
\end{table}

\section{Per-Baseline Integration Details}
\label{supp:integration}
\begin{figure}[t]
  \centering
    \includegraphics[width=\linewidth]{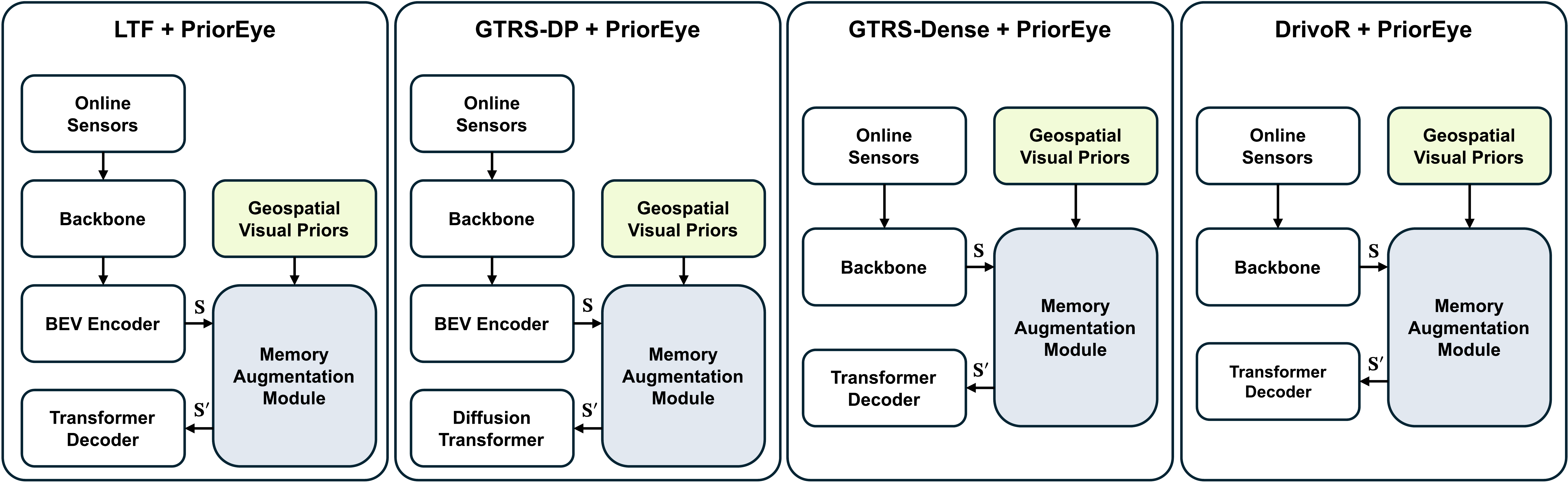}
    \caption{\textbf{Integration diagram for each baseline.}}
  \label{fig:integration_baseline}
\end{figure}

The memory augmentation module is identical across all four baselines: it augments the model-specific intermediate features with the retrieved geospatial visual priors as illustrated in \cref{fig:integration_baseline}. In every case, the intermediate state is formed by concatenating the perceptual features with the ego-status token. The specific inputs to the memory augmentation module for each baseline are summarized in Table~\ref{tab:integration_point}.

\begin{table}[ht]
  \centering
  \caption{\textbf{Integration point for each baseline.}}
  \label{tab:integration_point}
  \begin{NiceTabular}{l l}
    \toprule
    \textbf{Baseline} & \textbf{Intermediate features} \\
    \midrule
    LTF        & BEV features + ego-status \\
    GTRS-DP    & BEV features + ego-status \\
    GTRS-Dense & Image features + ego-status \\
    DrivoR     & Scene tokens + ego-status \\
    \bottomrule
  \end{NiceTabular}
\end{table}

\section{Practicality for Real-World Deployment}
\label{supp:deployment}
Most autonomous driving systems already maintain navigation paths for global routing. We believe PriorEye can be integrated into such systems without additional mapping effort: street-level imagery can be collected from their own fleet driving data, embedded, and anchored to their existing global routing map. Since the routing system already provides the intended route, the corresponding geospatial visual priors can be retrieved along that route and fed into the deployed end-to-end model.

\section{Comparison with HD Map Priors}
\label{supp:hdmap}

\begin{figure}[!t]
  \centering
  \includegraphics[width=\linewidth]{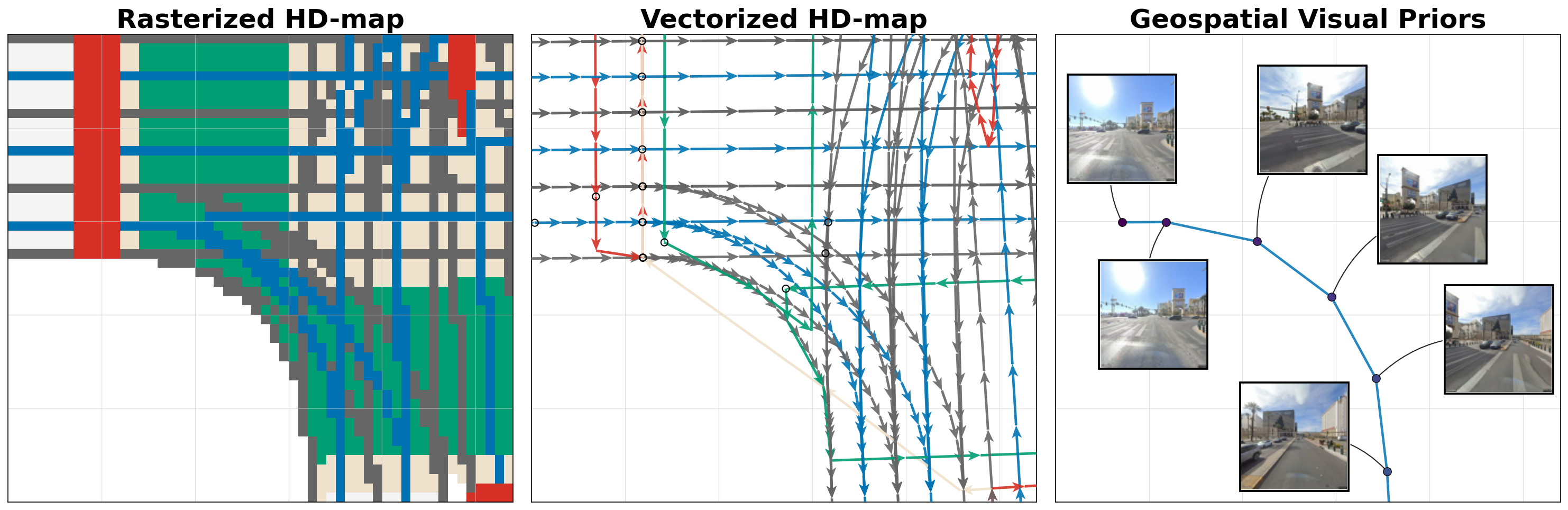}
  \caption{\textbf{Comparison of prior representations in the same scene.} From left to right: rasterized HD map, vectorized HD map, and geospatial visual priors. While the HD map variants encode a scene with rich map elements, our method uses street-view images and lane centers of the intended route.}
  \label{fig:prior_comparison}
\end{figure}

PriorEye anchors street-level visual priors to lane centers along the intended route. A natural question is whether comparable improvements could be achieved using HD maps, which encode rich structured priors. To investigate this, we compare PriorEye against two HD map baselines that follow standard map-encoding paradigms. The rasterized variant~\cite{Chauffeurnet} renders a 6-channel bird's-eye-view (drivable area, lane center, lane boundary, intersection, crosswalk, and stop line) at $0.5\,\mathrm{m/px}$, encoded by a strided CNN. The vectorized variant~\cite{Jiang_2023_ICCV, Gao_2020_CVPR} encodes five polyline types (excluding drivable area). Both variants additionally carry ego-relative location information that mirrors PriorEye's anchoring scheme: the rasterized variant through 2D positional embeddings, and the vectorized variant inherently through its ego-frame polylines. These baselines have access to richer map content than PriorEye, which uses only lane centers as illustrated in \cref{fig:prior_comparison}.

\begin{table}[ht!]
  \centering
  \setlength{\tabcolsep}{10pt}
  \caption{\textbf{Performance by prior type on \texttt{navhard-two-stage} (EPDMS).} The baseline is GTRS-Dense. HD map variants and PriorEye share an identical pipeline, differing only in the type of prior injected.}
  \label{tab:prior_map}
  \begin{NiceTabular}{c c c c}
    \toprule
    Baseline & + Vec. HD Map & + Rast. HD Map & \textbf{+ PriorEye} \\
    \midrule
    44.9 & 46.4\,{\scriptsize (+1.5)} & 47.1\,{\scriptsize (+2.2)} & \textbf{48.6\,{\scriptsize (+3.7)}} \\
    \bottomrule
  \end{NiceTabular}
\end{table}

As shown in \cref{tab:prior_map}, both HD map variants improve over the baseline, confirming that HD map content carries a useful signal. However, PriorEye yields the largest improvement (+3.7 EPDMS) while relying only on lane-center anchoring, a far lighter map requirement than full HD maps. Two observations follow. First, despite using only lane centers, PriorEye outperforms baselines with access to richer HD map content, indicating that the gain comes from the visual appearance along the route rather than from structured map elements. Second, geospatial visual priors require only encoding driving images along navigation lane centers, avoiding the construction and maintenance cost of a full HD map. Together, these results show that street-level visual priors are both more effective and practical than HD map priors for end-to-end autonomous driving.

\section{Input Ablation at Inference}
\label{supp:input_ablation}
To better understand how PriorEye uses its inputs and how sensitive it is to them, we analyze the behavior of the final trained model by varying its inputs at inference time, without any retraining. Unlike the ablations in the main paper, where each configuration is trained separately, here we fix the trained model and modify only its inference-time inputs. All results use GTRS-Dense as the baseline and are reported on \texttt{navhard-two-stage}.

\paragraph{Number of retrieved priors ($N$).}
We vary the number of retrieved priors at inference in Table~\ref{tab:n_scaling}. Performance peaks at $N{=}10$ (50\,m lookahead). We use $N{=}20$ in the main experiments, motivated by the $\sim$100\,m lookahead typical in autonomous driving. NAVSIM's scenarios are predominantly urban and low-speed, where such a long lookahead is unnecessary, making $N{=}10$ already sufficient.

\begin{table}[ht]
  \centering
  \footnotesize
  \setlength{\tabcolsep}{12pt}
  \caption{\textbf{Effect of the number of retrieved priors $N$ (EPDMS).}}
  \label{tab:n_scaling}
  \begin{NiceTabular}{c c c c c}
    \toprule
    $N$ & 5 & 10 & 15 & 20 \\
    \midrule
    EPDMS & 48.0 & 49.5 & 49.2 & 48.6 \\
    \bottomrule
  \end{NiceTabular}
\end{table}

\paragraph{Role of priors and perception.}
We probe the contribution of each input by removing it at inference time. When the model is evaluated without priors, EPDMS is 45.5, slightly above the baseline of 44.9, indicating that the persistent memory retains some benefit even without retrieved priors. When evaluated without perception, EPDMS drops sharply to 22.0, confirming that onboard perception remains essential and that the priors serve a complementary rather than a standalone role.

\section{Per-Metric Analysis}
\label{supp:permetric}

\begin{table}[ht]
  \centering
  \footnotesize
  \caption{\textbf{Per-metric comparison on the Stage~1 tokens of \texttt{navhard-two-stage} (450 tokens).} The baseline is GTRS-Dense. For each EPDMS sub-metric, we report the mean score of the baseline and PriorEye, their difference, and the number of tokens where each method wins (excluding ties).}
  \label{tab:per_metric}
  \renewcommand{\arraystretch}{1.0}
  \begin{NiceTabular}{l c c c c c}
    \toprule
    \textbf{Metric} & \textbf{Baseline} & \textbf{PriorEye} & \textbf{$\Delta$} & \textbf{Baseline wins} & \textbf{PriorEye wins} \\
    \midrule
    NC   & 98.4 & 97.6 & $-0.9$  & 7   & 3   \\
    DAC  & 95.8 & 97.1 & $+1.3$  & 12  & 18  \\
    DDC  & 99.4 & 100.0 & $+0.6$ & 0   & 3   \\
    TLC  & 99.3 & 99.8 & $+0.4$  & 0   & 2   \\
    EP   & 72.8 & 76.1 & $+3.3$  & 141 & 189 \\
    TTC  & 98.9 & 97.8 & $-1.1$  & 7   & 2   \\
    LK   & 94.7 & 97.3 & $+2.7$  & 2   & 14  \\
    HC   & 96.7 & 97.8 & $+1.1$  & 0   & 5   \\
    EC   & 40.4 & 52.0 & $+11.6$ & 42  & 94  \\
    \midrule
    \textbf{EPDMS} & \textbf{76.9} & \textbf{80.8} & $\mathbf{+3.9}$ & \textbf{143} & \textbf{230} \\
    \bottomrule
  \end{NiceTabular}
\end{table}

\begin{figure}[t]
  \centering
  \includegraphics[width=\linewidth]{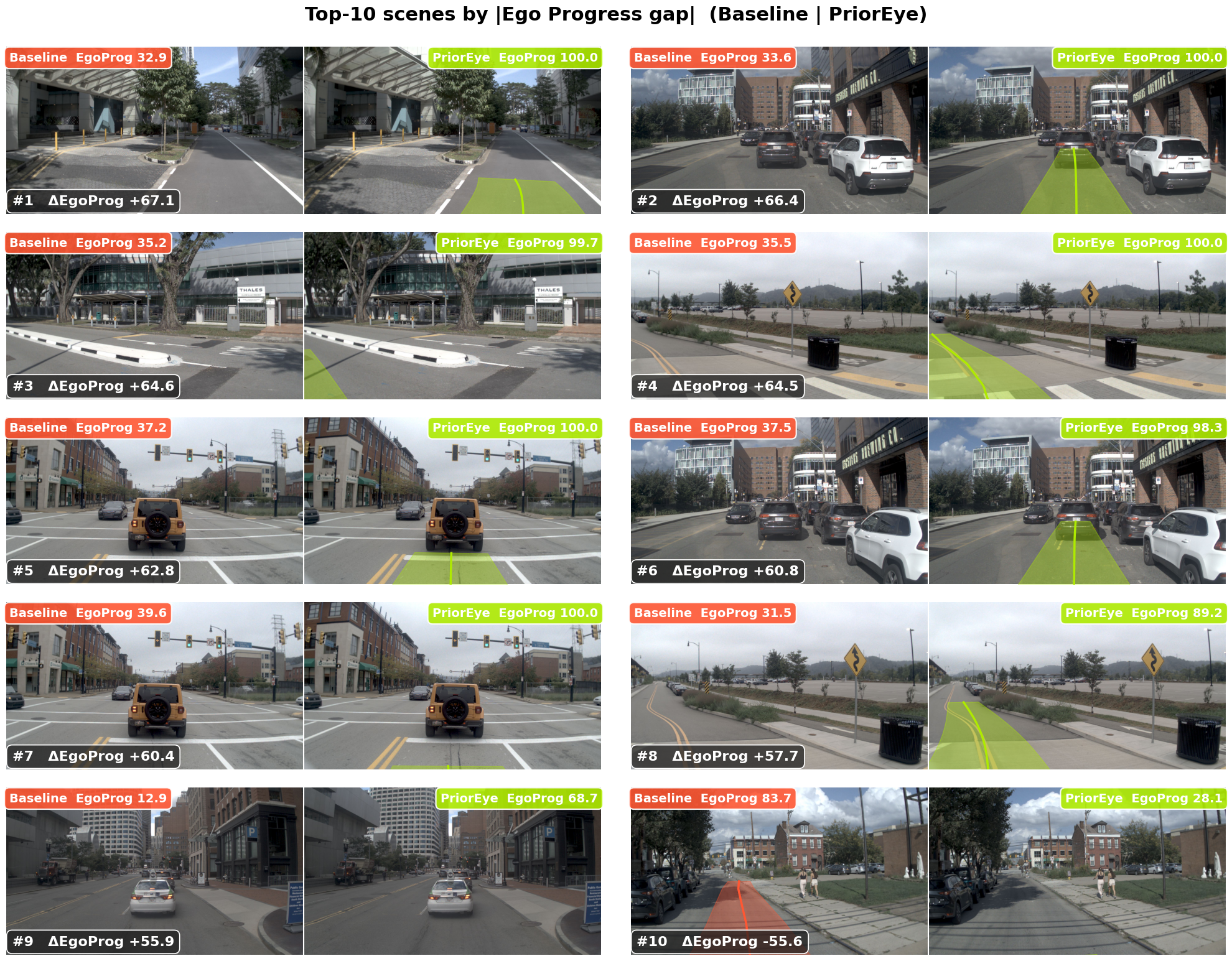}
  \caption{\textbf{Top-10 scenes with the largest absolute ego-progress difference on the Stage~1 tokens of \texttt{navhard-two-stage}, using GTRS-Dense as the baseline} (cf.\ Table~\ref{tab:per_metric}). For each scene, the baseline (left) and PriorEye (right) are shown with their ego-progress scores.}
  \label{fig:egoprog}
\end{figure}

Here we expand on the main results by analyzing the per-metric behavior of PriorEye in more detail. While the main paper reports the overall sub-metric scores, Table~\ref{tab:per_metric} additionally breaks down, for the real Stage~1 tokens, how often each method wins or loses on each metric, revealing where the improvement of PriorEye comes from. PriorEye improves the overall EPDMS, with gains spanning route adherence (DAC, DDC, LK), progress (EP), and comfort (HC, EC). The only metrics that slightly decline are the collision scores (NC, TTC). This pattern aligns with our intuition. The priors encode environmental context captured at a different time. Because road structure is largely static, priors directly help the model adhere to the route (DAC, DDC, LK) regardless of when they were collected. The same priors also give the model reliable foresight of the upcoming road, benefiting progress and comfort. A human driver navigates a familiar route more confidently and smoothly than an unfamiliar one. Likewise, knowing the road ahead lets the model avoid unnecessary hesitation, improving EP. By anticipating road features rather than reacting to them abruptly as they appear, the model produces smoother motion and better comfort (HC, EC). The same property explains the slight decline on the collision metrics (NC, TTC). The priors do not capture the current state of dynamic agents. They thus offer no benefit for avoiding surrounding agents, and the model's more confident progress occasionally reduces the safety margin to dynamic agents.

We further examine ego progress, the metric on which the two methods most often disagree, through qualitative examples. Figure~\ref{fig:egoprog} shows the ten scenes with the largest absolute difference in ego progress. In nine of them (\#1 to \#9), PriorEye progresses substantially more than the baseline. In most of these scenes, the path ahead is occluded, either by other vehicles (\#2, \#5, \#6, \#7, \#9) or by road geometry such as curves (\#3, \#4). The retrieved priors reveal the upcoming route, allowing the model to avoid unnecessary hesitation. In the remaining scene (\#10), the baseline shows higher ego progress, but it crosses the road boundary.

\section{Qualitative Results for Robustness Test}
\label{sec:supp_robustness}
\subsection{Sensor Robustness}
\label{sec:supp_sensor_robustness}
\begin{figure}[!t]
  \centering
  \includegraphics[width=0.9\linewidth]{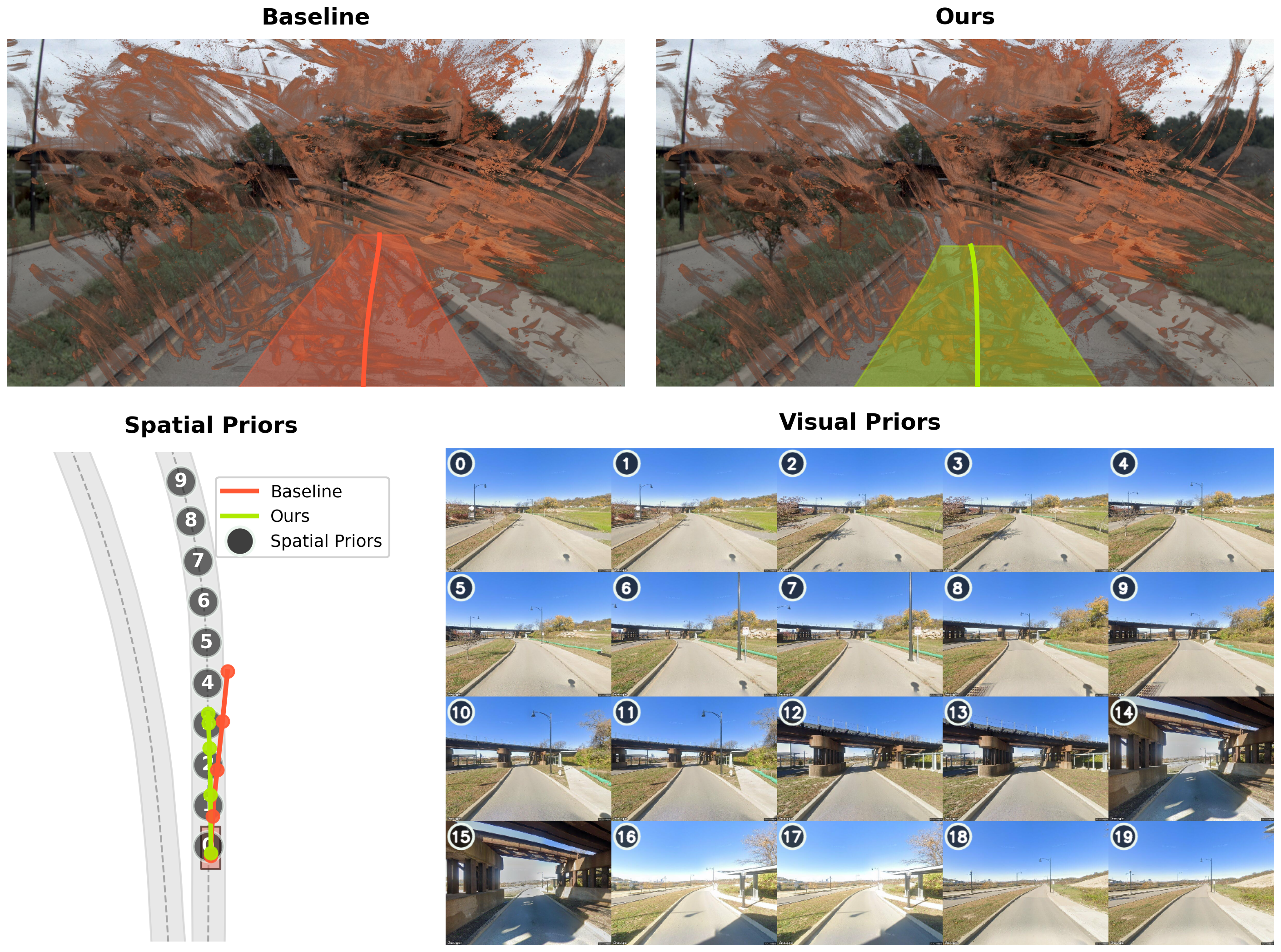}
  \caption{\textbf{Qualitative comparison under sensor degradation (Mud (Heavy))}. The baseline (GTRS-Dense, left) generates a trajectory that deviates toward the road boundary, while our method (GTRS-Dense + PriorEye, right) plans a safe, lane-keeping trajectory. The retrieved geospatial visual priors are unaffected by the sensor corruption, providing reliable scene context despite the degraded onboard cameras.}
  \label{fig:sensor_robustness}
\end{figure}

Fig.~\ref{fig:sensor_robustness} shows a scenario where the onboard cameras are heavily occluded by mud splatter. While the baseline fails to plan a safe trajectory under such degradation, the proposed method maintains proper lane-keeping. This is because the geospatial visual priors are captured independently of the onboard sensors and thus remain unaffected by real-time corruption, providing clean scene context that compensates for the degraded observations.

\subsection{Geospatial Visual Prior Corruption}
\label{sec:supp_corruption}
\begin{figure}[!t]
  \centering
  \begin{subfigure}[b]{\textwidth}
    \centering
    \includegraphics[width=0.95\linewidth]{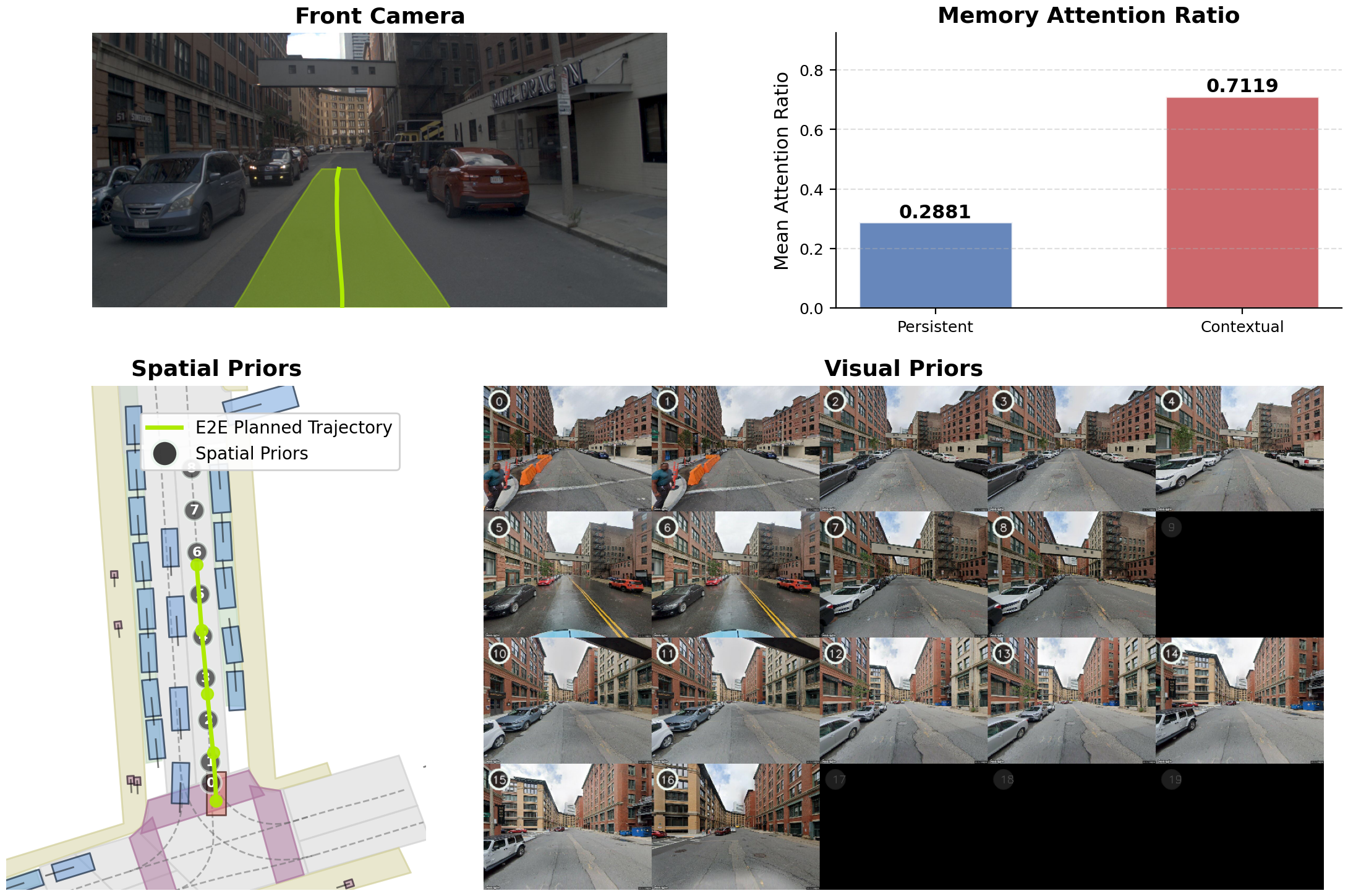}
    \caption{Normal retrieval.}
    \label{fig:corruption_normal}
  \end{subfigure}\\[6pt]
  \begin{subfigure}[b]{\textwidth}
    \centering
    \includegraphics[width=0.95\linewidth]{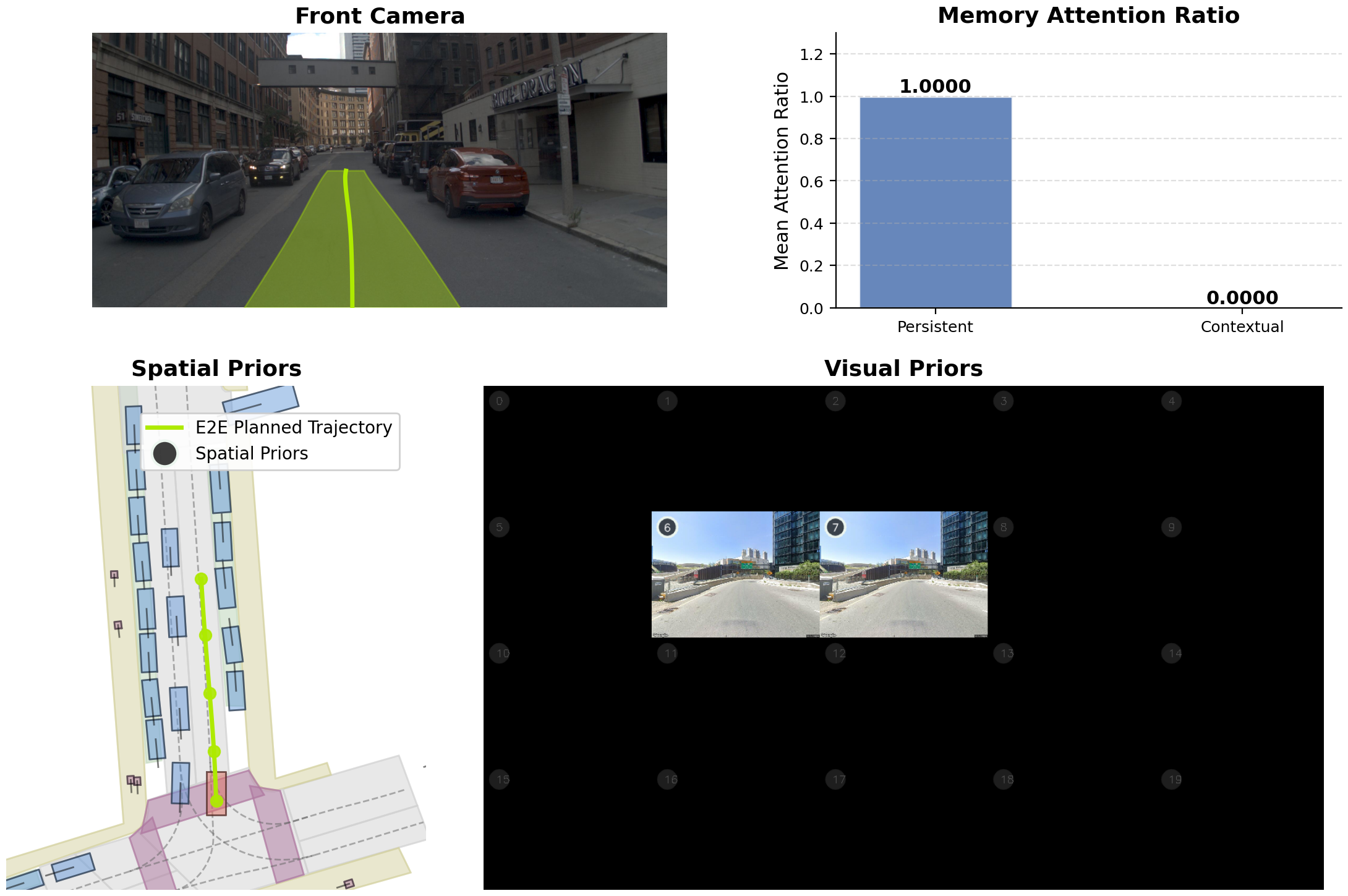}
    \caption{Corrupted retrieval (${\sim}500$\,m offset).}
    \label{fig:corruption_corrupted}
  \end{subfigure}
  \caption{\textbf{Qualitative comparison under geospatial visual prior corruption}. (a)~Under normal operation, contextual memory dominates attention (0.71). (b)~When priors are retrieved from ${\sim}500$\,m away, the proposed module assigns full attention to persistent memory (1.00), ignoring the irrelevant context. The planned trajectory remains similar in both cases, demonstrating graceful degradation under the corrupted-prior condition.}
  \label{fig:corruption}
\end{figure}

Fig.~\ref{fig:corruption} shows a scenario where the geospatial retrieval query is perturbed, causing priors to be fetched from approximately 500\,m away. While most retrieved priors become empty (black images) or irrelevant under such corruption, the planned trajectory remains largely consistent with that of normal retrieval. This is because the proposed method adaptively shifts its attention entirely to persistent memory ($1.00$), effectively discarding the unreliable contextual input.

\end{document}